\documentclass[a4paper, 10pt]{article}
\usepackage{graphicx}
\usepackage[font=footnotesize]{caption}
\usepackage{subcaption}
\usepackage{amsmath}
\usepackage[numbers,sort&compress]{natbib}
\usepackage{hyperref}
\usepackage{mathtools}
\usepackage{authblk}
\usepackage{amsfonts}
\usepackage{amssymb}
\usepackage{float}
\usepackage{booktabs}
\usepackage{tikz-cd}
\usepackage{multirow}
\usepackage{times}
\usepackage[capitalize,nameinlink]{cleveref}
\usepackage[margin=1in]{geometry}

\title{How Molecular Generative Models Organize Molecular Identity}

\author[1]{Raul Ortega-Ochoa\thanks{Corresponding author: raul.ortega-ochoa@tri.global, mail.raulortega@gmail.com}}
\author[2,3]{Tejs Vegge}
\author[1]{Jens S. Bakander}
\author[4,5,10]{Luis Mantilla Calderón}
\author[4,5,6,7,8,9,10]{Alán Aspuru-Guzik}
\author[11]{Tonio Buonassisi}

\affil[1]{Toyota Research Institute, Los Altos, California, USA}
\affil[2]{Department of Energy Conversion and Storage, Technical University of Denmark}
\affil[3]{CAPeX Pioneer Center for Accelerating P2X Materials Discovery, Kgs. Lyngby, Denmark}
\affil[4]{Department of Computer Science, University of Toronto, Toronto, ON, Canada}
\affil[5]{Vector Institute for Artificial Intelligence, Schwartz Reisman Innovation Campus, Toronto, ON, Canada}
\affil[6]{Department of Chemistry, University of Toronto, Toronto, ON, Canada}
\affil[7]{Department of Chemical Engineering \& Applied Chemistry, University of Toronto, Toronto, ON, Canada}
\affil[8]{Department of Materials Science \& Engineering, University of Toronto, Toronto, ON, Canada}
\affil[9]{Acceleration Consortium, Toronto, ON, Canada}
\affil[10]{NVIDIA, 431 King St. W \#6th, Toronto, ON, Canada}
\affil[11]{Department of Mechanical Engineering, Massachusetts Institute of Technology, Cambridge, MA, USA}
\date{}

\begin{document}

\maketitle
\vspace{-2.2em}

\begin{abstract}
Generative models for matter are often evaluated as samplers over output representations, and their latent spaces are commonly used as proxies for navigating chemical space. Much less is known about how these models internally arrange discrete chemical identities within those representations. We study this arrangement by making molecular identity explicit and pulling it back through the generative process. Through these pullbacks we probe the regions that generate the same object, exposing the trained model’s internal repertoire: a fixed partition that determines which objects (novel or not) the model can produce.

Across three molecular generative architectures, we find that this repertoire is arranged into piecewise-constant regions separated by recurring coarse-to-fine boundaries. Its organization depends on the representation probed, the identity convention, decoder stochasticity, and the metric used to compare coordinates. During training, local chemical organization stabilizes while the number of distinct molecular identities represented within each neighborhood continues to change. Internal organization must therefore be characterized, rather than assumed, before a generative space can be treated as chemically navigable.
\end{abstract}

\section{Introduction}

Evaluations of generative models emphasize validity, novelty, diversity, likelihood, and downstream utility. These measures characterize what a model produces, but leave a complementary structural question: how are the identities it can produce arranged within the generative process? Where is molecular identity resolved? Do nearby coordinates decode to chemically related molecules? How many distinct identities occupy a neighborhood, and how does this organization emerge during decoding and training? Answering these questions first requires specifying what counts as the same object. Generative models produce representations, but the represented objects of interest are defined only after specifying which representations count as equivalent.

Molecular generation offers a favorable testbed for this question. Under a stated convention such as canonical SMILES~\cite{smiles,cansmiles} or InChI~\cite{Heller2015}, strings, graphs, and other representations can be compared reproducibly for molecular identity. These conventions still encode choices about stereochemistry, tautomerism, and protonation state, but they let us distinguish the generated representation from the molecular object it denotes.

Continuous molecular representations have supported interpolation, optimization, and inverse design \cite{G_mez_Bombarelli_2018,doi:10.1126/science.aat2663}. Subsequent work has clarified complementary aspects of this navigation~\cite{schiff2020characterizing}: structured molecular decoders improve validity~\cite{Jin2018JunctionTV}, contrastive objectives encourage alignment between coordinate distance and molecular structure~\cite{kirchoff2023salsa}. Together, these studies sharpen different aspects of what is often called latent-space smoothness or navigability. A complementary object remains less directly formalized: once a molecular identity convention is chosen, how are the resulting discrete identities arranged across the coordinates and stochastic choices of a trained generator? This connects to studies of decision-region geometry and convexity~\cite{Tetkova2025} and to broader geometric views of conceptual spaces~\cite{10.7551/mitpress/2076.001.0001}. Here, however, the labels are not fixed dataset classes but equivalence classes of representations.

We make molecular identity explicit as an equivalence relation on the output representation space and pull the resulting identity classes back through the generative process. For stochastic generators, we expose decoding randomness as a `random tape' $\eta$, so that a pair \((z,\eta)\) determines an output. This induces a partition of \(\mathcal Z \times \mathcal{E}\) by molecular identity. The partition exists by construction; its sectional structure, chemical cohesiveness, persistence across decoder randomness, alignment with coordinate metrics, and evolution during training are empirical properties. 

We study these properties in MolMiner~\cite{ortegaochoa2025molminercontrollable3dawarefragmentbased}, a property-conditioned autoregressive model; HierVAE~\cite{jin2020hierarchicalgenerationmoleculargraphs}, a hierarchical graph variational autoencoder; and GDSS~\cite{pmlr-v162-jo22a}, a score-based graph diffusion model. Fixed-randomness sections of all three models exhibit piecewise-constant identity regions. When the random tape changes, MolMiner and HierVAE retain chemically cohesive local neighborhoods, whereas GDSS shows little exact-identity persistence and clearer organization only under coarser identity conventions. HierVAE further shows that local chemical cohesiveness need not be globally ordered by Euclidean or cosine relationships between coordinates. In MolMiner, intermediate decoding states support sequential trajectory divergence as one mechanism for the observed coarse-to-fine boundaries. During training, chemical cohesiveness stabilizes before the number of observed identities within a neighborhood finishes evolving.

A continuous generative coordinate space is therefore not sufficient to establish a navigable chemical space. Its internal organization must be characterized relative to the representation being generated, the convention used to define molecular identity, the stochastic decoding process, and the metric used to compare coordinates.

\section{Theory \& Definitions}
\label{sec:theory}

Let \(\mathcal X\) denote the output representation space of molecules (e.g., strings, graphs, point clouds in 3D space). An identity convention is an equivalence relation \(x\sim y\) on \(\mathcal X\). The choice of the convention, which declares when two representations are equivalent, is domain-dependent. What counts as the same object depends on a convention of what object identity is, and is external to the representation space.

Within a representation space \(\mathcal X\), the represented object (e.g., a molecule) is not merely a single representation \(x\in\mathcal X\), but the set of all representations equivalent under a particular convention `$\sim$', the equivalence class \([x]_\sim=\{y\in\mathcal X:y\sim x\}\). Applying an adopted convention of \emph{equivalence} `$\sim$' through representation space \(\mathcal X\) creates molecular space, a quotient space \(\mathcal X/\!\sim=\{[x]_\sim:x\in\mathcal X\}\). 

In this work we use six conventions of molecular equivalence. In each case $x_1\sim x_2$ iff the two representations agree on the named invariant: \textbf{SMILES}: canonical isomeric SMILES~\cite{smiles,cansmiles}. Preserves atom counts, bond orders, aromaticity, and stereochemistry. \textbf{InChIKey-14}: First 14-character block of the InChIKey~\cite{Heller2015}, a hash of the connectivity layer. Collapses stereoisomers, some tautomers (via InChI normalization), and protonation states. \textbf{Formula}: Molecular formula, e.g.\ $\mathrm{C_9H_8O_4}$. \textbf{Elements}: Set of heavy-atom element symbols, e.g.\ $\{\mathrm{C,O}\}$; no counts or positions. Coarser than formula. \textbf{Murcko}: Bemis--Murcko scaffold~\cite{doi:10.1021/jm9602928}: iteratively remove every atom not in a ring and not on a ring--ring path. \textbf{Murcko-generic}: The Murcko scaffold with every atom set to carbon and every bond to single, keeping only the ring/linker skeleton.

We differentiate a molecule \([x]_\sim\) from its representation \(x\). While a representation space can be continuous (e.g., point clouds in 3D space), molecular space, for example under `\(\sim\,\leftrightarrow\,\)same InChIKey-14', is discrete. Finally, to preempt the question of what molecule an invalid representation (e.g., a failed-syntax SMILES string) corresponds to, we include a `null molecule' \(\varnothing\in\mathcal X/\!\sim\).

\paragraph{Controlling the random-tape.} Generators are often stochastic. The same internal vector \(z\in\mathcal Z\) may decode to different outputs under different sampling choices. To treat them deterministically, we expose all stochasticity through a random tape \(\eta\in\mathcal E\), so that once \((z,\eta)\in\mathcal{Z}\times \mathcal{E}\) is fixed, the forward map is deterministic\footnote{For deterministic generators, \(\mathcal E\) is a singleton.}. While we are not able to navigate the random tape of a decoder, which has no obvious metric, we can freeze random tapes \(\eta\in\mathcal E\) to study which part of the \((z,\eta)\)-axis dominates in determining the decoded molecule and its internal chemical neighborhood. In other words, we use it to study first whether there is chemical coherence for a neighborhood of $z$ at constant $\eta$, and only then we study whether this chemical organization in $z$ persists for variable $\eta$.

\paragraph{Internal coordinates index decoding paths.}
The internal vector \((z,\eta)\) indexes a computational trajectory in the model. The generator produces a representation $x$ through a sequence of intermediate states $\{s_i\in\mathcal{S}\}$, a computational path $\tau\in\mathcal P$, resulting in a molecular representation $x\in\mathcal X$.
What plays the role of a state \(s_i\) is architecture-dependent: a partial molecule in an autoregressive generator, a partially denoised representation in a diffusion model, or a layerwise hidden representation in a feedforward neural network.

\paragraph{Tracking the molecule back to the source.}
The forward map from \((z,\eta)\in\mathcal{Z}\times \mathcal{E}\) coordinates to molecules admits a backwards one-to-many reading, tracking back along its chain all the successive pre-images: from the representation class of the molecule, to the trajectories that create them, to the set of coordinates that source them, $C_{[x]_\sim}$. Eq.~\ref{eq:pullback-diagram} illustrates the forward map and the fibers of the corresponding pullback.

\begin{equation}
\begin{tikzcd}[row sep=2.5em, column sep=3em]
\mathcal{Z} \times \mathcal{E} \arrow[r, "\Gamma_\theta"] 
  & \mathcal{P} \arrow[r, "D"] 
  & \mathcal{X} \arrow[r, "q_\sim"] 
  & \mathcal{X}/\!\sim \\
C_{[x]_\sim} \arrow[u, dashed, no head] 
  & T_{[x]_\sim} \arrow[u, dashed, no head] \arrow[l, "\Gamma_\theta^{-1}"'] 
  & {\{y\in\mathcal{X}:y\sim x\}} \arrow[u, dashed, no head] \arrow[l, "D^{-1}"'] 
  & {[x]_\sim} \arrow[u, dashed, no head] \arrow[l, "q_\sim^{-1}"']
\end{tikzcd}
\label{eq:pullback-diagram}
\end{equation}

These pullback components collect, respectively, all output representations of the molecule, all generation paths that decode it, and all coordinates that generate it. The probability of generating a molecule \([x]_\sim\) can then be written as mass over the cell. Repetition, memorization, novelty, and local search of molecules can be studied through the geometry and mass allocation of these cells.
\begin{equation}
p_\theta([x]_\sim)
= \int_{C_{[x]_\sim}} \rho(z,\eta)\,dz\,d\eta , \qquad \text{with}\qquad C_{[x]_\sim} = \{(z,\eta)\in\mathcal Z\times\mathcal E : D(\Gamma_\theta(z,\eta))\sim x\}.
\label{eq:mass-equivalence}
\end{equation}

The cell picture naturally invites the question of cell volume. The geometric volume of a cell depends on a chosen geometry on $\mathcal{Z}\times\mathcal{E}$. Native-coordinate volume is one possible convention, but it has no automatic chemical meaning. A task-relevant notion of volume would instead require, much like identity, an external metric to be pulled back through the generator. Such decoder-induced geometries are generally position-dependent and need not be flat~\cite{arvanitidis2018latent}.

\paragraph{Cells are piecewise-constant regions.} While the \emph{cell partition} induced by the pullback on \(\mathcal{Z} \times \mathcal{E}\) exists by construction, its organization and boundary structure are properties to be tested. We hypothesize the partition is formed by identity regions whose sections form piecewise-constant patches. Two arguments make this plausible: First, networks with piecewise-linear activations partition their input space into a polyhedral complex of regions on which the network is affine~\cite{Montfar2014OnTN,pmlr-v80-balestriero18b}. In this sense, our hypothesized partition is a coarsening of the decoder's polyhedral decomposition under a given molecular equivalence relation. Second, viewing the decoder as assigning each \(z\) to an identity class in \(\mathcal X/\!\sim\) is analogous to a classifier assigning inputs to classes, and recent empirical work suggests that decision regions in neural-network representations tend toward convexity across architectures and domains~\cite{Tetkova2025}.

\section{Experiments}
\label{sec:experiments}

We probe the internal organization on \(\mathcal{Z} \times \mathcal{E}\) of three different molecular generative architectures: MolMiner~\cite{ortegaochoa2025molminercontrollable3dawarefragmentbased}, a decoder-only autoregressive transformer whose \(\mathcal{Z}\) is a vector of physicochemical properties, HierVAE~\cite{jin2020hierarchicalgenerationmoleculargraphs}, a hierarchical graph variational autoencoder~\cite{Kingma2014} with an autoregressive decoder, and GDSS~\cite{pmlr-v162-jo22a}, a score-based diffusion model over molecular graphs where \(\mathcal{Z}\) is the prior-noise. All three models are trained on ZINC~\cite{doi:10.1021/ci3001277}: for GDSS we use the official checkpoints; for HierVAE and MolMiner we retrain in-house to access intermediate training checkpoints.

\subsection{Varying coordinates $z\in \mathcal{Z}$ at constant random tape $\eta \in \mathcal{E}$}
\label{subsec:deterministic-probe}
We study the partition directly in its native coordinates. Since \(\mathcal Z\times\mathcal E\) is high-dimensional and dimensionality reduction would introduce geometric artifacts and distortions, we instead probe different lower dimensional cross-sections of this high-dimensional space while holding the random tape constant by fixing the seed. 

If cells are cohesive territories tiling up the space, we expect decoded molecules to remain the same over intervals and switch abruptly at cell boundaries. We probe this both along one-dimensional straight-line paths interpolating between two samples \(z_0, z_1\), \(\alpha\in[0,1]\mapsto z'=(1-\alpha)z_0+\alpha z_1\) (SI Section~\ref{sec:si-1d-paths}), and in two-dimensional sections. For MolMiner, the primary section varies the first two coordinates of the conditioning vector, corresponding to logP~\cite{Wildman1999PredictionOP} and QED~\cite{Bickerton2012}, while holding the remaining coordinates fixed. For HierVAE, we vary a fixed pair of latent dimensions while holding all other dimensions fixed. For GDSS, we vary a fixed pair of scalar entries of the prior-noise tensor, in either the atom-feature block \(z_x\) or the adjacency block \(z_{\mathrm{adj}}\), holding the rest fixed. In all cases, the random seed is fixed, and alternative coordinate-pair sections are reported in SI Section~\ref{sec:si-alt-slices}.

\subsection{Varying the random tape $\eta \in \mathcal{E}$ at constant coordinates $z\in \mathcal{Z}$}
\label{subsec:stochastic-probe}
After the analysis of the partition on \(\mathcal{Z}\) for fixed-\(\eta\), we study the complementary question of whether the same molecule can be persistently decoded for a fixed $z\in\mathcal{Z}$ along the random-tape axis $\mathcal E$. Along a straight-line path in \(\mathcal{Z}\), we run the decoder in its stochastic sampling mode, as opposed to the fixed-tape decoding of Sections~\ref{subsec:deterministic-probe} and~\ref{subsec:training-snapshots-probe}, decoding each intermediate point multiple times with different random seeds, and record the empirical distribution over canonical-SMILES identities at successive points along the path. For GDSS we additionally decode a single fixed coordinate \(z\) under independent noise realizations and record the empirical frequency distribution over canonical-SMILES identities.

\subsection{Chemical organization of $\mathcal Z$-neighborhoods of $\mathcal Z \times \mathcal E$}
\label{subsec:chemical-cohesiveness-probe}

We decode local neighborhoods by sampling within fixed-radius balls in \(\mathcal Z\) space, without fixing the random tape, effectively sampling over all \(\mathcal E\). In the following we refer to these balls as neighborhoods. Centered on a reference point \(z_0\), we define the ball
\(
\mathcal B(z_0,r) = \{z\in\mathcal Z : \|z-z_0\|_2 \le r\},
\)
and draw \(N\) points uniformly from this ball using~\cite{HARMAN20102297}:
\[
z^{(k)} = z_0 + r\,(u^{(k)})^{1/d}\frac{v^{(k)}}{\|v^{(k)}\|}, \qquad v^{(k)}\sim\mathcal N(0,I_d), \quad u^{(k)}\sim\mathcal U(0,1).
\]
The centers \(\{z_0^{(i)}\}\) are stratified by log-density under the reference sampling density \(\rho(z)\) so that the across-neighborhood baseline samples meaningfully separated parts of \(\mathcal Z\). The radius is set to \(r=0.1\), with \(N=100{,}000\) decodes per ball. MolMiner and GDSS use \(30\) balls, HierVAE uses \(60\). Each sampled point is decoded once and labeled with the six identity conventions.

We compare the decoded molecules within the same neighborhood, or across different ones. We measure Jaccard~\cite{jaccard1901etude} set-overlap between the molecules decoded from these neighborhoods, using all six conventions of molecular equivalence to probe whether the same molecules can be decoded from distinct regions in \(\mathcal Z\). Then, we compute Tanimoto~\cite{tanimoto1958elementary} similarities on Extended-Connectivity fingerprints (ECFP)~\cite{doi:10.1021/c160017a018,doi:10.1021/ci100050t}, and MACCS~\cite{doi:10.1021/ci010132r}, to grade whether molecules within (\(W\)) chemical neighborhoods are more similar than across (\(A\)) different neighborhoods, aggregating the results through the area under the curve:
\[
\mathrm{AUC}(W,A) = \Pr\left(T_W > T_A\right),
\]
where \(T_W\) and \(T_A\) denote Tanimoto similarities sampled from the within- and across-neighborhood pairs. Values above \(0.5\) indicate that molecules decoded from the same neighborhood are more chemically similar than molecules decoded from different neighborhoods. The same neighborhood data is then used to test whether the median Tanimoto similarity on ECFP and MACCS fingerprints between two neighborhoods correlates with the Euclidean distance or cosine similarity between their centers.

\subsection{Training evolution of chemical organization of $\mathcal Z$-neighborhoods}
\label{subsec:training-snapshots-probe}

To study how the partition and its chemical organization change during training, we repeat the neighborhood analysis of Section~\ref{subsec:chemical-cohesiveness-probe} across a sequence of checkpoints. This time we hold the random tape fixed and look only at the $\mathcal Z$-axis, since over $\mathcal E$ the number of distinct molecules in a neighborhood keeps growing. To compensate, and to define neighborhoods large enough to contain sufficiently many distinct molecules for the statistical analysis, we increase the neighborhood radius from $r=0.1$ to $0.5$, reusing the same centers as in the previous analysis. At each checkpoint we draw \(N=2000\) samples per neighborhood recording (i)~the number of unique molecules (under SMILES convention); (ii)~within-ball versus across-ball ECFP Tanimoto measured through \(\mathrm{AUC}(W,A)\) and (iii)~cross-ball Jaccard overlap of decoded sets under the six equivalence conventions.

For a directly inspectable view of the same dynamics, we also track a single two-dimensional section (as in Section~\ref{subsec:deterministic-probe}) across checkpoints for MolMiner and HierVAE. At each checkpoint we decode the grid, label the molecules by canonical-SMILES, and record the number of distinct molecules.

\section{Results}
\label{sec:results}
\subsection{Fixing the random tape reveals the piecewise-constant regions}
\label{subsec:results-fixed-sections}

\begin{figure}[H]
    \centering
    \begin{subfigure}[b]{0.31\linewidth}
        \centering
        \includegraphics[width=\linewidth]{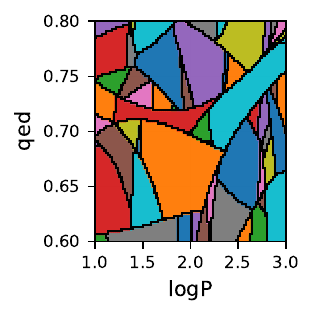}
        \caption{MolMiner}
        \label{fig:slice-molminer}
    \end{subfigure}
    \hfill
    \begin{subfigure}[b]{0.31\linewidth}
        \centering
        \includegraphics[width=\linewidth]{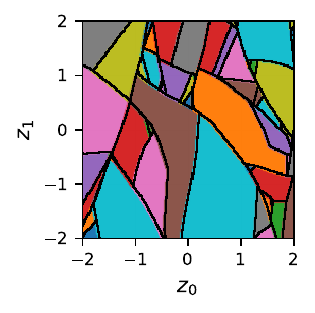}
        \caption{HierVAE}
        \label{fig:slice-hiervae}
    \end{subfigure}
    \hfill
    \begin{subfigure}[b]{0.31\linewidth}
        \centering
        \includegraphics[width=\linewidth]{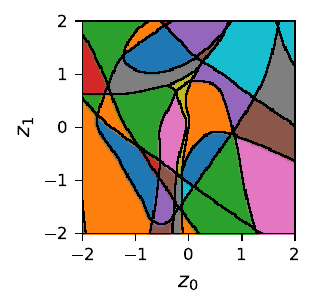}
        \caption{GDSS}
        \label{fig:slice-gdss}
    \end{subfigure}

    \caption{%
        \textbf{Two-dimensional fixed-\(\eta\) sections.}
        Each grid point is decoded with a fixed random seed and colored by canonical-SMILES identity. These maps visualize sections of the partition. For MolMiner, the section varies the first two conditioning coordinates, corresponding to logP and QED. Colors cycle within each panel and do not identify the same molecule across panels or models.
    }
    \label{fig:deterministic-tessellation}
\end{figure}

Figure~\ref{fig:deterministic-tessellation} examines the partition within two-dimensional fixed-\(\eta\) sections, showing contiguous patches across the three models. For each model we additionally decode five distinct two-dimensional sections at the same seed; the piecewise-constant identity-cell structure persists across all of them, confirming that it is not an artifact of the particular slice shown here (SI Section~\ref{sec:si-alt-slices}). Re-labeling the same slices under the five other conventions merges some of these regions into larger patches, highlighting the dependence of the resulting partition on the chosen equivalence (SI Section~\ref{sec:si-equiv-relations}).

The sections invite different interpretations across models. In MolMiner and HierVAE, nearby intervals along the path often preserve recognizable scaffold-level structure while making small chemical edits (see SI Fig.~\ref{fig:si-deterministic-tessellation} for annotated molecules along straight-line paths through \(\mathcal Z\)), suggesting that the visible patches are chemically organized; this is tested quantitatively in Section~\ref{subsec:results-chemical-cohesiveness}.

\subsection{Molecular persistence: same $z \in \mathcal{Z}$, different noise $\eta \in \mathcal E$}
\label{subsec:results-stochastic}

\begin{figure}[H]
    \centering
    \begin{subfigure}[t]{0.48\linewidth}
        \centering
        \includegraphics[width=\linewidth]{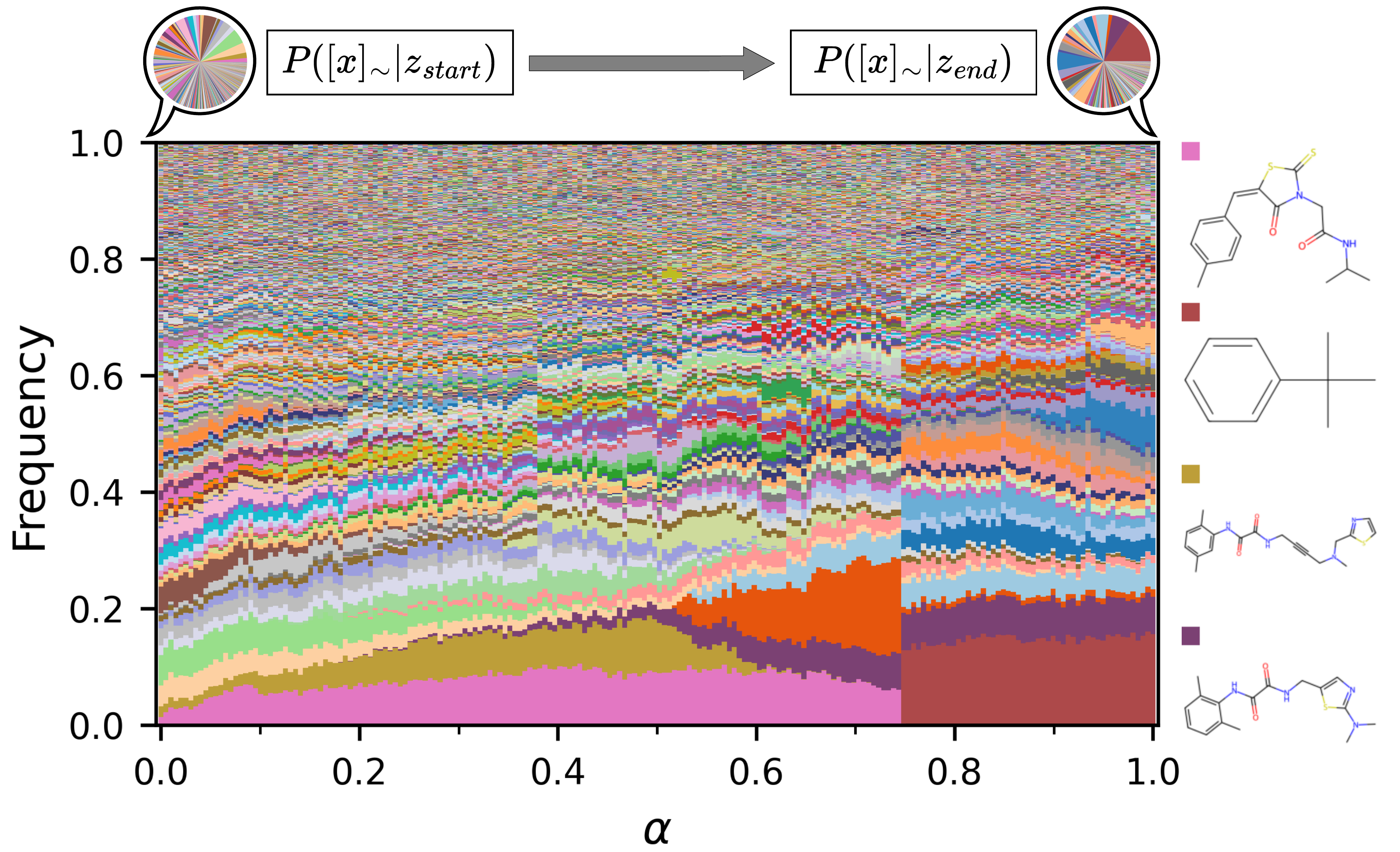}
        \caption{MolMiner}
        \label{fig:prob-decoding-molminer}
    \end{subfigure}
    \hfill
    \begin{subfigure}[t]{0.48\linewidth}
        \centering
        \includegraphics[width=\linewidth]{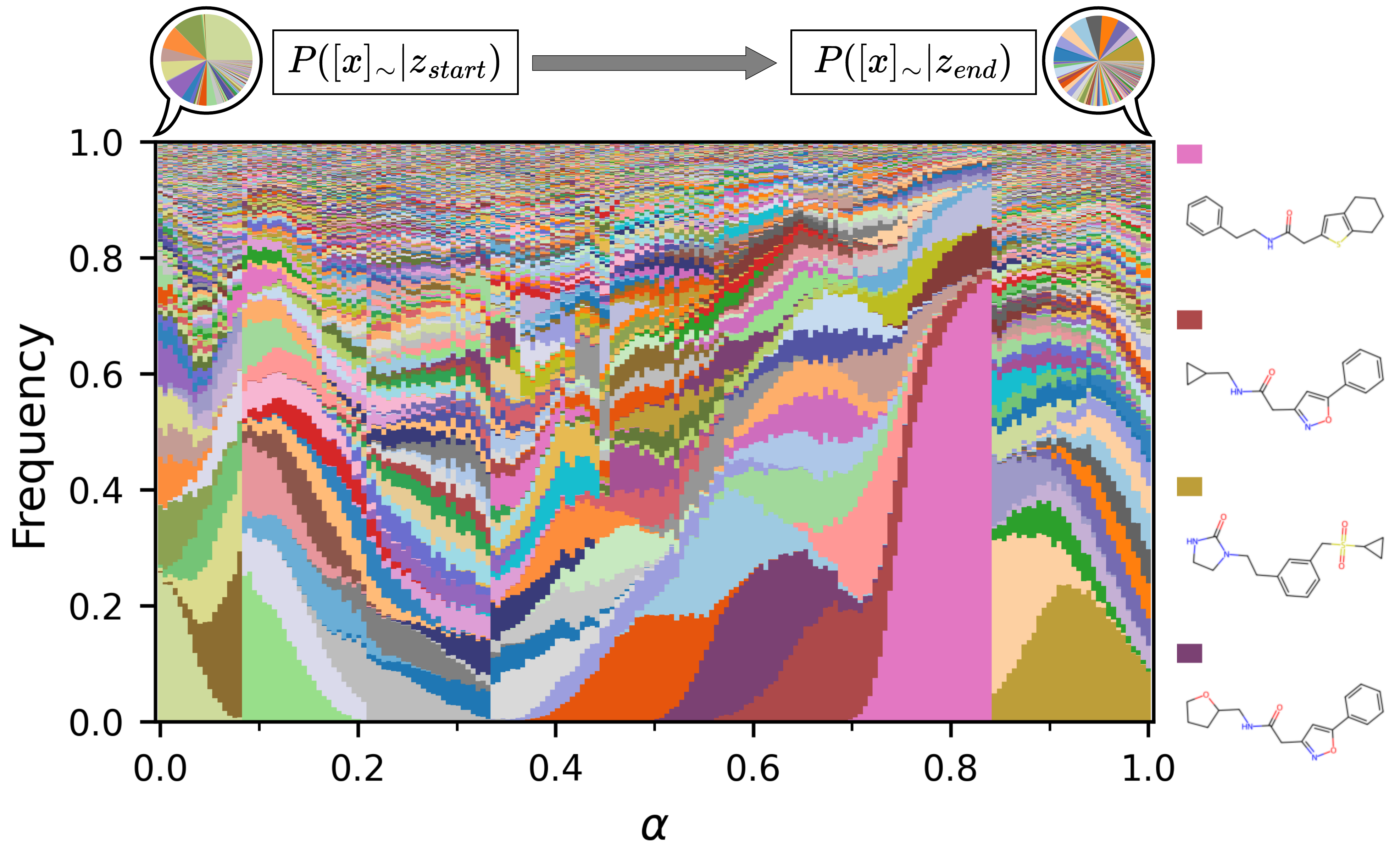}
        \caption{HierVAE}
        \label{fig:prob-decoding-hiervae}
    \end{subfigure}
    \caption{%
    \textbf{Dependence of decoded molecules along a path (stochastic).} Straight-line paths in \(\mathcal{Z}\) for MolMiner and HierVAE with \(\sim 5{,}000\) resamples per \(z\). At each of 200 intermediate points \(z\), the decoder is sampled repeatedly with varying random seed, resulting in a distribution over molecules. Segment heights reflect relative frequency. In both models, which molecules are decoded varies systematically along the path, indicating that \(z\) retains finer control over the molecules decoded, as opposed to GDSS.
    }
    \label{fig:prob-decoding}
\end{figure}

For MolMiner and HierVAE, Fig.~\ref{fig:prob-decoding} shows a straight-line path in \(\mathcal Z\). At each coordinate \(z\) along the path the decoder is sampled repeatedly in stochastic mode with different random seeds, and the resulting empirical distribution over canonical-SMILES is visualized as a flow-plot. In both models, the identity distribution evolves along the path, indicating that \(z\) retains control over the resulting molecule for varying \(\eta\). GDSS behaves differently: repeated decoding at a single fixed \(z\) produces almost entirely unique canonical SMILES, with \(102{,}400\) independent noise realizations resulting in \(102{,}348\) distinct canonical SMILES (\(99.95\%\) unique molecules), and the most frequent SMILES, pyridine, occurred only six times (\(0.006\%\)). Therefore, at SMILES resolution, the flow-plot is structureless (SI Figure~\ref{fig:flow-gdss-canonical}). 

Under coarser equivalence relations the flow plots become smoother and the number of distinct molecules under such conventions decreases (SI~\ref{sec:si-equiv-relations}, Figures~\ref{fig:si-flow-hiervae-equiv}--\ref{fig:si-flow-gdss-equiv}). For GDSS, under the InChIKey-14 and formula conventions, the flow-plot remains structureless (SI, Figures~\ref{fig:flow-gdss-inchikey}--\ref{fig:flow-gdss-formula}). However, structure appears under the element, Murcko, and Murcko-generic conventions (SI, Figures~\ref{fig:flow-gdss-composition}--\ref{fig:flow-gdss-murcko-generic}). GDSS therefore appears organized only at coarse resolution, weakly conditioned on \(z\) at scaffold or element resolution, consistent with its greater Jaccard overlap than MolMiner and HierVAE at the coarse equivalence relations of Table~\ref{tab:jaccard-summary}.

\subsection{Neighborhoods are chemically organized in MolMiner and HierVAE}
\label{subsec:results-chemical-cohesiveness}

We report median across-neighborhood Jaccard overlap under each of the six conventions described in Section~\ref{sec:theory} with the results presented in Table~\ref{tab:jaccard-summary}. Because overlaps are computed between sets from \emph{different} neighborhoods, low values mean distinct neighborhoods occupy distinct regions of chemistry and high values mean they converge on common classes. Overlap tends to be larger on coarser relations, so the informative quantity is the resolution at which overlap is appreciable. For MolMiner and HierVAE it stays near zero through SMILES, InChIKey-14, Murcko, and formula, becoming appreciable (0.29 and 0.28) only at bare element: distinct neighborhoods occupy genuinely distinct coarse-chemical regions and coincide only once chemistry is reduced to its constituent elements. GDSS behaves differently: overlap is negligible at SMILES and InChIKey-14 levels (\(5.6\times10^{-5}\) and \(6.2\times10^{-5}\)) but rises steeply once the quotient coarsens, to 0.21 under formula and 0.71 under element, far above MolMiner and HierVAE at the same resolutions. Relative to those two models, GDSS shows noticeably less neighborhood-localized chemical structure: its identity sets reuse shared coarse chemistry across neighborhoods rather than staying confined to distinct territories.

\begin{table}[H]
\centering
\caption{
\textbf{Molecule overlap across neighborhoods.}
Median Jaccard indices between molecule sets decoded from different neighborhoods.}
\label{tab:jaccard-summary}
\setlength{\tabcolsep}{6pt}
\renewcommand{\arraystretch}{1.2}
\begin{tabular}{lcccccc}
\toprule
Architecture & SMILES & InChIKey-14 & Murcko & Murcko-generic & Formula & Elements  \\
\midrule
MolMiner & $0$                  & $0$                  & $1.5\mathrm{e}{-3}$ & $2.4\mathrm{e}{-2}$ & $1.4\mathrm{e}{-3}$ & $0.29$  \\
HierVAE  & $0$                  & $0$                  & $1.3\mathrm{e}{-3}$ & $2.0\mathrm{e}{-2}$ & $2.0\mathrm{e}{-3}$ & $0.28$  \\
GDSS     & $5.6\mathrm{e}{-5}$  & $6.2\mathrm{e}{-5}$  & $1.9\mathrm{e}{-2}$ & $8.7\mathrm{e}{-2}$ & $0.21$              & $0.71$  \\
\bottomrule
\end{tabular}
\end{table}

\begin{figure}[h]
    \centering
    \includegraphics[width=\linewidth]{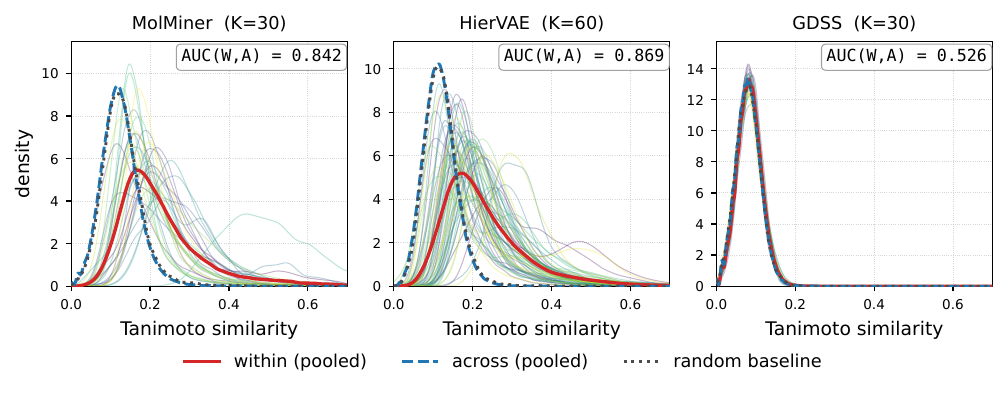}
    \caption{Distributions of pairwise Tanimoto similarity on ECFP fingerprints for within-neighborhood (\(W\)), across-neighborhood (\(A\)), and random baseline (\(R\)) pairs of molecules. In MolMiner and HierVAE, \(W\) is shifted toward higher similarity than \(A\) and \(R\), indicating chemical organization at the neighborhood scale; GDSS shows near-coincident \(W\)/\(A\)/\(R\) distributions.}
    \label{fig:chemical-cohesiveness-kde}
\end{figure}

We compute Tanimoto similarities on ECFP and MACCS fingerprints between molecule pairs drawn from the same neighborhood (within, W), from different neighborhoods (across, A), and from the full pool at random (baseline, R), and ask whether two molecules from the same neighborhood are more chemically similar than two from different neighborhoods. MolMiner and HierVAE show chemical organization under this probe (Figure~\ref{fig:chemical-cohesiveness-kde}). In MolMiner, within-neighborhood molecules have median Tanimoto similarity \(0.203\), compared with \(0.121\) for across and \(0.125\) for the random baseline. The pooled probability that a within-neighborhood pair is more similar than an across-neighborhood pair is \(\mathrm{AUC}(W,A)=0.842\), with \(\mathrm{AUC}(W,R)=0.825\) as control. For HierVAE within-neighborhood median Tanimoto similarity is \(0.203\), versus \(0.115\) across and \(0.117\) random baseline with \(\mathrm{AUC}(W,A)=0.869\) and \(\mathrm{AUC}(W,R)=0.859\) as control. For GDSS within-neighborhood median Tanimoto similarity is \(0.083\), compared with \(0.080\) for across and random pairs. The pooled \(\mathrm{AUC}(W,A)=0.526\) is only slightly above chance, and the absolute separation between \(W\), \(A\), and \(R\) is small.

The combination of negligible exact-molecule overlap and chemical differentiation within neighborhoods indicates that the partition in MolMiner and HierVAE is chemically organized: nearby coordinates decode to chemically related molecules, and different neighborhoods occupy distinct regions of chemistry.

\subsection{Euclidean distance and cosine similarity do not reliably track chemical organization}
\label{subsec:results-metric-diagnostic}
Given that the partition of MolMiner and HierVAE is chemically organized, a natural follow-up question is whether Euclidean distance and cosine similarity in these partition coordinates are faithful proxies for chemical distance. This question is implicit in any practice that uses these distances for interpolation, local search, novelty radii, or latent navigation generally. We test it directly by regressing median pairwise ECFP Tanimoto similarity between neighborhoods on the Euclidean distance and, separately, the cosine similarity between their centers.

For MolMiner, Euclidean distance between neighborhood centers partially predicts median ECFP Tanimoto similarity (\(R^{2}_{\mathrm{euc}}=0.33\)). This is largely built in: the conditioning vector is physicochemical descriptors, so Euclidean distance in conditioning space approximates a physicochemical metric. The non-trivial part is that the decoder preserves it: decoded molecules still satisfy the imposed organization. Cosine similarity tracks this chemical distance poorly (\(R^{2}_{\mathrm{cos}}=0.12\)). For GDSS, with no organization to capture (\(\mathrm{AUC}(W,A)=0.526\)), the regression has nothing to track in this setting, and its flatness (\(R^{2}_{\mathrm{euc}}\approx R^{2}_{\mathrm{cos}}\approx0\)) is consistent with that. In HierVAE the two probes disagree: cohesiveness is strong (\(\mathrm{AUC}(W,A)=0.869\)). Yet Euclidean distance and cosine similarity between neighborhood centers explain almost none of the variance in chemical similarity (\(R^{2}_{\mathrm{euc}}=0.05\); \(R^{2}_{\mathrm{cos}}=0.05\)). We read this as a metric diagnostic: the latent space is chemically organized at the local scale probed by decoded balls, but Euclidean distance or cosine similarity between centers does not reliably preserve chemical proximity across neighborhoods. This is consistent with broader evidence that deep generative latent spaces are often curved rather than flat~\cite{arvanitidis2018latent}. For downstream use, the practical implication is that Euclidean radii, interpolation lengths, and nearest-neighbor searches in HierVAE-like latent spaces should not automatically be interpreted as chemical distances. 

Table~\ref{tab:cohesion-summary} consolidates the comparison. Reading across columns separates two questions: whether the identity-cell partition is chemically organized (the AUC columns) and whether Euclidean distance or cosine similarity are faithful proxies for that organization (the \(R^2\) and \(\rho\) columns). To confirm the picture is not specific to ECFP, we recomputed all diagnostics on MACCS~\cite{doi:10.1021/ci010132r} keys and found the same architecture-dependent ordering: MolMiner and HierVAE strongly cohesive, GDSS near chance. The metric relationship weakens further under MACCS (MolMiner \(R^2_{\mathrm{euc}}\) falls from \(0.33\) to \(0.13\); HierVAE remains near zero).

\begin{table}[H]
\centering
\caption{
\textbf{Cross-architecture summary of chemical organization and metric scaling, across fingerprints.}
Medians are pairwise Tanimoto similarities for within- (\(W\)), across- (\(A\)), and random (\(R\)) neighborhood pairs, computed on ECFP (Morgan, \(r{=}2\), 2048-bit) or MACCS (166-bit) fingerprints. \(\mathrm{AUC}(W,A)\) is the probability that a randomly chosen within-neighborhood pair has higher Tanimoto similarity than a randomly chosen across-neighborhood pair (\(0.5\) is chance). \(R^2\) and Spearman \(\rho\) summarize the relationship between Euclidean distance or cosine similarity in generative coordinates and median pairwise chemical similarity between neighborhoods. Sub/superscripts denote the interquartile range. Neighborhoods per architecture: \(30\) for MolMiner and GDSS, \(60\) for HierVAE.
}
\label{tab:cohesion-summary}
\setlength{\tabcolsep}{4pt}
\renewcommand{\arraystretch}{1.3}
\begin{tabular}{llccccccccc}
\toprule
Fingerprint & Architecture & med\(W\) & med\(A\) & med\(R\) & AUC\(_{W,A}\) & AUC\(_{W,R}\) & \(R^{2}_{\mathrm{euc}}\) & \(\rho_{\mathrm{euc}}\) & \(R^{2}_{\mathrm{cos}}\) & \(\rho_{\mathrm{cos}}\) \\
\midrule
\multirow{3}{*}{ECFP}
 & MolMiner & $0.20^{0.27}_{0.16}$ & $0.12^{0.15}_{0.09}$ & $0.13^{0.16}_{0.10}$ & $0.84^{0.91}_{0.78}$ & $0.83^{0.91}_{0.76}$ & 0.33 & $-0.51$ & 0.12 & $+0.34$ \\[2pt]
 & HierVAE  & $0.20^{0.27}_{0.16}$ & $0.11^{0.14}_{0.09}$ & $0.12^{0.14}_{0.09}$ & $0.87^{0.95}_{0.83}$ & $0.86^{0.95}_{0.81}$ & 0.05 & $-0.19$ & 0.05 & $+0.21$ \\[2pt]
 & GDSS     & $0.08^{0.10}_{0.06}$ & $0.08^{0.10}_{0.06}$ & $0.08^{0.10}_{0.06}$ & $0.52^{0.53}_{0.52}$ & $0.52^{0.54}_{0.52}$ & 0.00 & $-0.00$ & 0.00 & $-0.05$ \\
\midrule
\multirow{3}{*}{MACCS}
 & MolMiner & $0.56^{0.68}_{0.45}$ & $0.37^{0.45}_{0.29}$ & $0.39^{0.48}_{0.31}$ & $0.82^{0.93}_{0.78}$ & $0.78^{0.91}_{0.73}$ & 0.13 & $-0.36$ & 0.05 & $+0.21$ \\[2pt]
 & HierVAE  & $0.54^{0.64}_{0.44}$ & $0.38^{0.46}_{0.31}$ & $0.39^{0.47}_{0.32}$ & $0.81^{0.90}_{0.77}$ & $0.80^{0.90}_{0.72}$ & 0.02 & $-0.15$ & 0.04 & $+0.19$ \\[2pt]
 & GDSS     & $0.41^{0.48}_{0.33}$ & $0.39^{0.47}_{0.31}$ & $0.39^{0.47}_{0.32}$ & $0.54^{0.57}_{0.52}$ & $0.54^{0.59}_{0.49}$ & 0.02 & $+0.14$ & 0.00 & $-0.05$ \\
\bottomrule
\end{tabular}
\end{table}

\subsection{Testing a sequential origin for coarse-to-fine boundaries}
\label{subsec:results-branching}

Across all three models, the fixed-randomness sections in Figure~\ref{fig:deterministic-tessellation} exhibit a recurring coarse-to-fine pattern: broad molecular territories separated by major boundaries contain smaller regions separated by finer boundaries. The final maps alone cannot establish whether this organization is genuinely hierarchical or explain how it arises. A natural hypothesis is that it reflects the sequential structure of decoding. Coordinates whose trajectories share early intermediate states initially belong to the same broad region, and boundaries appear when those trajectories diverge. Divergences early in decoding would then create coarse boundaries, whereas later divergences would subdivide the resulting regions more finely.

MolMiner allows us to test this hypothesis directly because every intermediate decoding state is a valid partial molecular graph. We decode the same two-dimensional section with a fixed random seed and record the partial molecule at every generation step. At each step, the grid can therefore be labeled by its current partial structure rather than only by its final molecular identity. If the coarse-to-fine appearance of Figure~\ref{fig:deterministic-tessellation} arises through sequential branching, early decoding steps should divide the section into a small number of broad regions, later boundaries should appear primarily within those regions, and the major boundaries between early lineages should remain visible in the final map.

\begin{figure}[H]
    \centering
    \begin{subfigure}{\linewidth}
        \includegraphics[width=\linewidth]{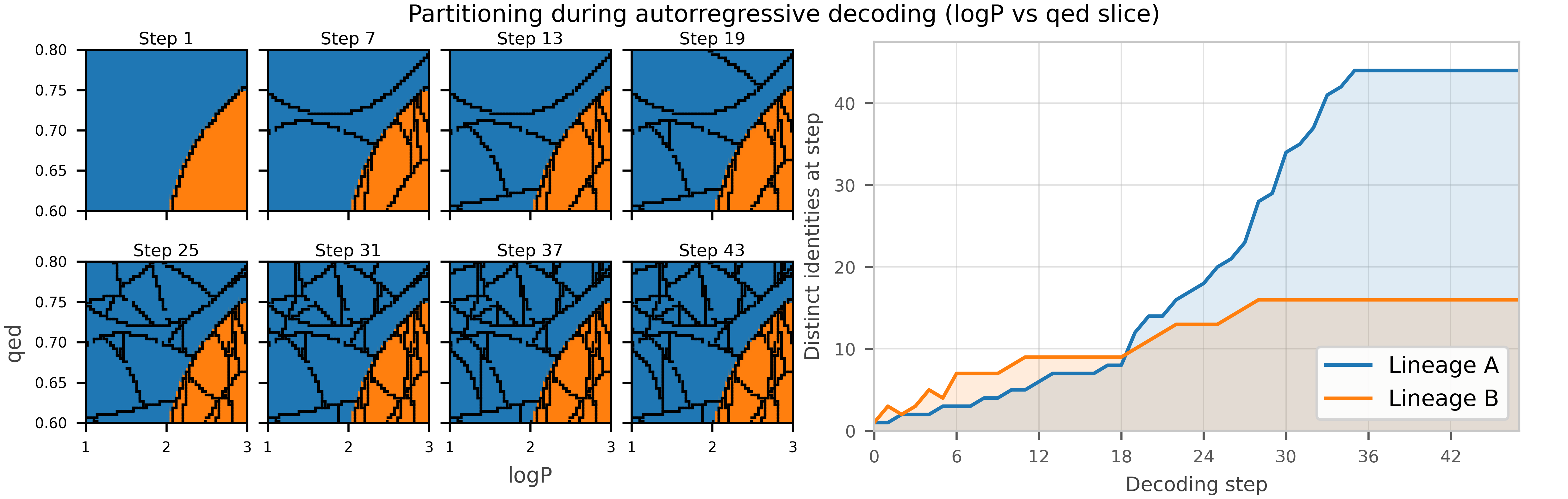}
    \end{subfigure}
    \begin{subfigure}{\linewidth}
        \includegraphics[width=\linewidth]{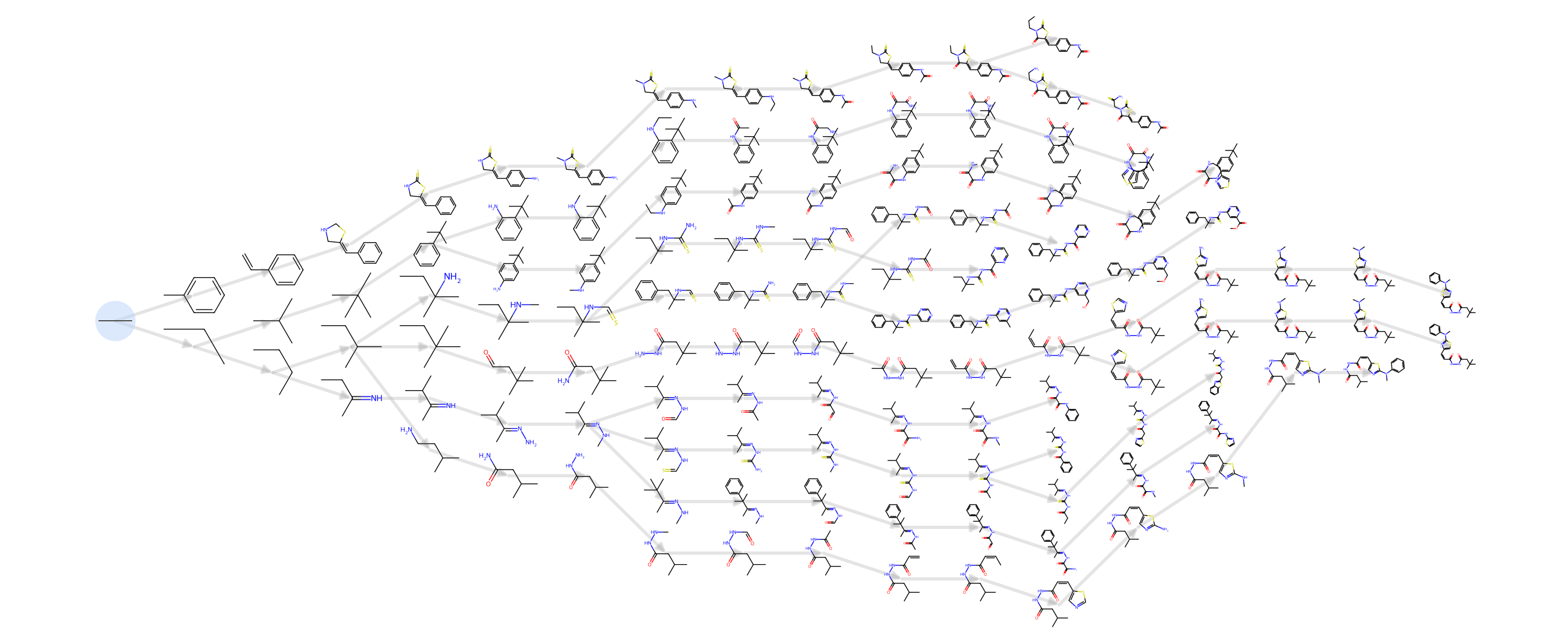}
    \end{subfigure}
    \caption{
    \textbf{Partition subdivision as a function of decoding depth.}
    Partition of a fixed MolMiner two-dimensional section shown as a function of decoding step. Blue and orange label cells originating from two distinct starting fragments (lineages). New boundaries appear as subdivisions within each lineage while coarser inter-lineage boundaries remain largely intact, consistent with final molecular regions being formed through sequential subdivision of shared partial trajectories. Top right: number of distinct partial-molecule states as a function of decoding step. Bottom: compact representation of Lineage B showing distinct intermediate structures.
    }
    \label{fig:lineages}
\end{figure}

Figure~\ref{fig:lineages} shows the resulting sequence. Early decoding steps divide the section into broad regions associated with different initial fragments; the blue and orange regions mark two such lineages. As decoding proceeds, new boundaries appear predominantly within each lineage, while the coarse boundary separating them remains largely intact. The final molecular regions are therefore formed progressively: points that share an early partial structure remain grouped until their decoding trajectories diverge, after which the corresponding region is subdivided. An independent section shows the same coarse-to-fine pattern in SI Figure~\ref{fig:si-lineages}.

The curves in the top-right panel summarize the increasing granularity of the two lineages. The number of distinct partial-molecule states generally rises with decoding depth as additional branching decisions separate previously shared trajectories. This evolution is not a strict refinement at every step. Occasional decreases, particularly early in Lineage B, show that trajectories distinguished at one stage can later reconsolidate into the same molecular identity. Such reconsolidation is possible because MolMiner is order-agnostic: the same final molecule can be constructed through different fragment-addition orders. A direct example of distinct construction histories reaching the same final molecular identity is shown in SI Figure~\ref{fig:si-traj}. The bottom panel of Figure~\ref{fig:lineages} provides a compact trace of the successive subdivisions and reconsolidations within Lineage B.

The coarse-to-fine appearance of the HierVAE and GDSS sections is consistent with an analogous process in which decoding trajectories diverge from shared intermediate states. The present experiment, however, demonstrates this mechanism only for MolMiner. Applying the same analysis to HierVAE would require exposing and aligning its intermediate hierarchical decoding states across the coordinate section. For GDSS, the difficulty is more fundamental: intermediate reverse-time states are noisy graphs that frequently do not correspond to valid molecules, so the molecular identity labels used for the final sections cannot be applied at each denoising step. The similar final patterns in these models therefore motivate the same hypothesis, but do not by themselves establish a common mechanism.

\subsection{Chemical organization stabilizes before molecular granularity}
\label{subsec:results-training}

Having examined how molecular boundaries emerge during decoding, we next ask how the local organization of generated molecules develops during training. We repeat the matched-neighborhood analysis of Section~\ref{subsec:training-snapshots-probe} at a sequence of training checkpoints. The same neighborhood centers and decoding protocol are used throughout, with the random tape fixed, so changes in the resulting statistics reflect changes in the learned coordinate-to-molecule map.

For MolMiner we use epochs 1--50 and the best-validation checkpoint; because its decoder has no termination cap, in early checkpoints (epochs $\sim$1--3) some neighborhoods contain samples resembling polymers (Fig.~\ref{fig:runaway} in the SI). These polymer-like structures, which we refer to as runaways, stem from the untrained autoregressive policy recurrently attaching the same motif indefinitely, making these neighborhoods prohibitively expensive to sample. Our analysis is therefore restricted to a subset of \(20\) balls that complete at every checkpoint. For HierVAE we use steps 5k--240k; its decoder enforces a decode cap at~\(150\) steps, so no runaway occurs at early checkpoints.

We track two properties of each neighborhood. Chemical cohesiveness measures whether molecules decoded from the same neighborhood are more similar than molecules decoded from different neighborhoods, summarized by \(\mathrm{AUC}(W,A)\). Molecular granularity is the number of distinct molecular identities decoded within a neighborhood. These quantities capture different aspects of organization: cohesiveness measures how chemically localized a neighborhood is, whereas granularity measures how finely the decoder resolves that neighborhood into distinct molecules.

\begin{figure}[H]
    \centering
    \begin{subfigure}{\linewidth}
        \centering
        \includegraphics[width=\linewidth]{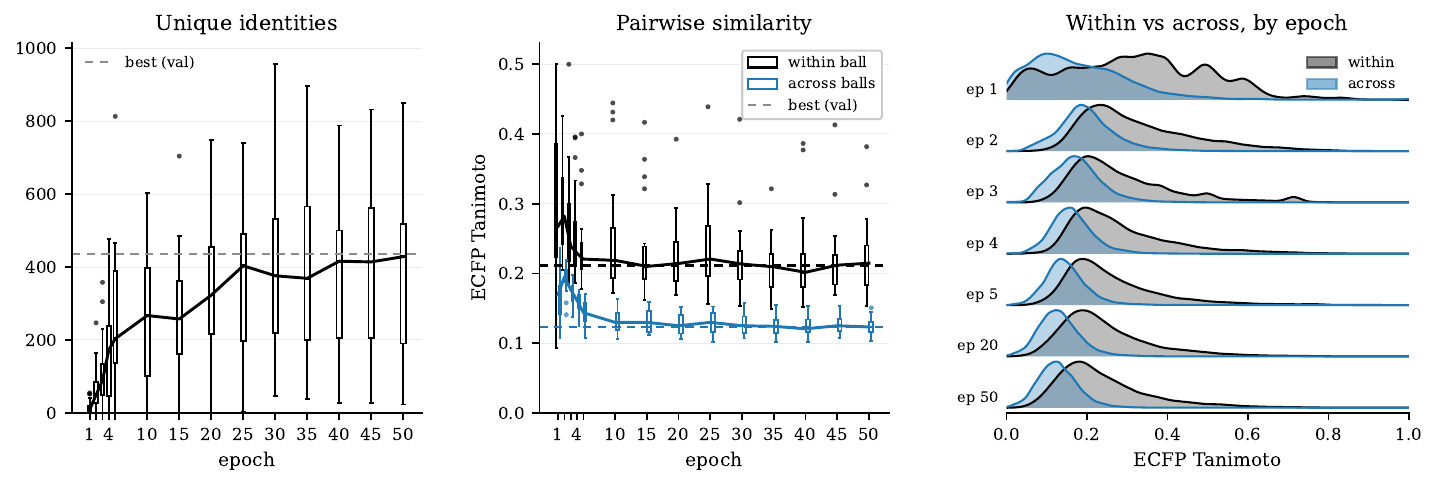}
        \caption{MolMiner's internal organization evolution.}
    \end{subfigure}

    \begin{subfigure}{\linewidth}
        \centering
        \includegraphics[width=\linewidth]{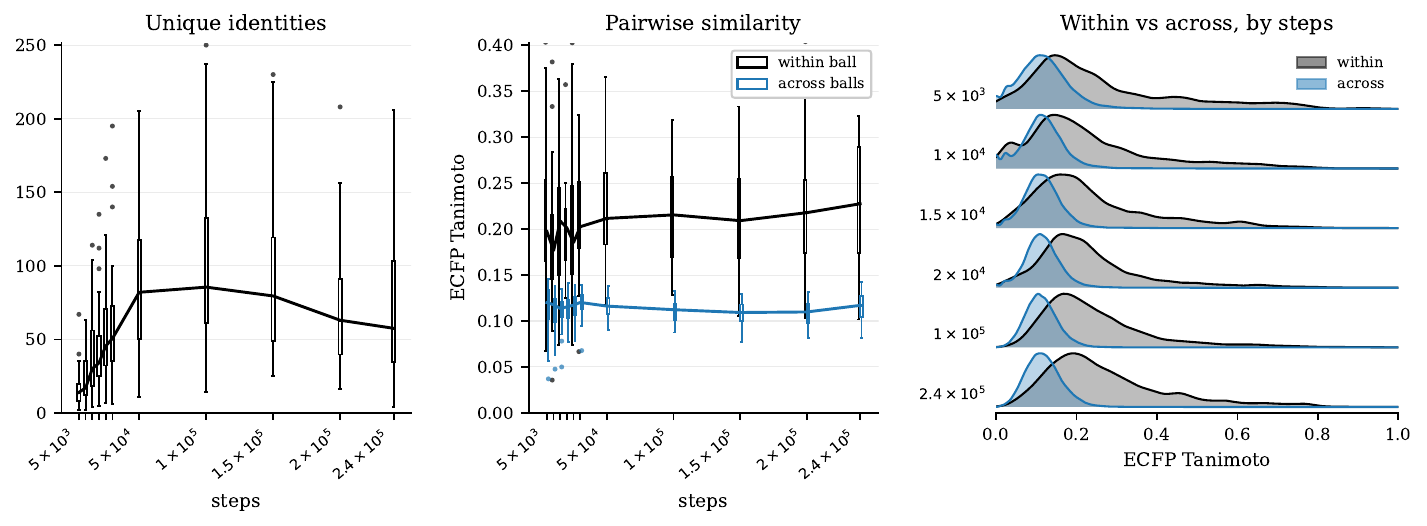}
        \caption{HierVAE's internal organization evolution.}
    \end{subfigure}

    \caption{\textbf{Training-time evolution of local chemical organization and molecular granularity.} Statistics computed at coordinate neighborhoods ($r=0.5$, $N=2000$ deterministic decodes/neighborhood) across training checkpoints. \textbf{(a)} MolMiner across epochs. \textbf{(b)} HierVAE across optimization steps. \emph{Left:} number of unique molecular identities per ball (box: interquartile range over balls), measuring local identity granularity. \emph{Center:} median within- ($W$) versus across-neighborhood ($A$) ECFP Tanimoto similarity; the persistent gap between $W$ and $A$ is the cohesiveness signal. \emph{Right:} densities of within- and across-neighborhoods Tanimoto similarity at selected checkpoints, shown as stacked ridgelines. Chemical cohesiveness stabilizes before the number of identities per neighborhood finishes evolving.}
    \label{fig:evolution-convergence}
\end{figure}

Both models show a clear separation between the timescales of chemical cohesiveness and molecular granularity. From Figure~\ref{fig:evolution-convergence}, cohesiveness increases early and then remains approximately stable: by about epoch~5 in MolMiner and \(50\)k optimization steps in HierVAE. The number of distinct identities, in contrast, continues to change after cohesiveness has plateaued. In MolMiner, granularity continues to increase and begins to level off only toward the end of training. In HierVAE, it rises rapidly, reaches a maximum near \(100\)k steps, and then contracts.

For MolMiner, the median number of unique identities per neighborhood increases from \(9\) at epoch~1 to approximately \(430\) near the end of training, while \(\mathrm{AUC}(W,A)\) rises from \(0.74\) to a stable range of approximately \(0.83\)--\(0.86\). Cross-neighborhood overlap at the element level remains between \(0.27\) and \(0.38\) from epoch~5 onward, suggesting that the increase in cohesiveness is not driven by a large change in the coarse elemental-composition overlap between neighborhoods. HierVAE shows the same qualitative decoupling through a different granularity trajectory: the median number of identities rises from approximately \(14\) at \(5\)k steps to a peak of approximately \(86\) near \(100\)k steps, before decreasing to approximately \(58\)--\(63\). Its \(\mathrm{AUC}(W,A)\) increases from \(0.78\) to a stable range of approximately \(0.87\)--\(0.88\).

The within-neighborhood Tanimoto distributions lie to the right of the across-neighborhood distributions in both models, with most of the separation developing early in training (Fig.~\ref{fig:evolution-convergence}, right). Together, these results suggest that broad local chemical cohesiveness is established before the decoder finishes refining the number of distinct molecules represented within each neighborhood. Increasing molecular granularity therefore does not require a corresponding continued increase in cohesiveness.

\begin{figure}[H]
    \centering
    \includegraphics[width=\linewidth]{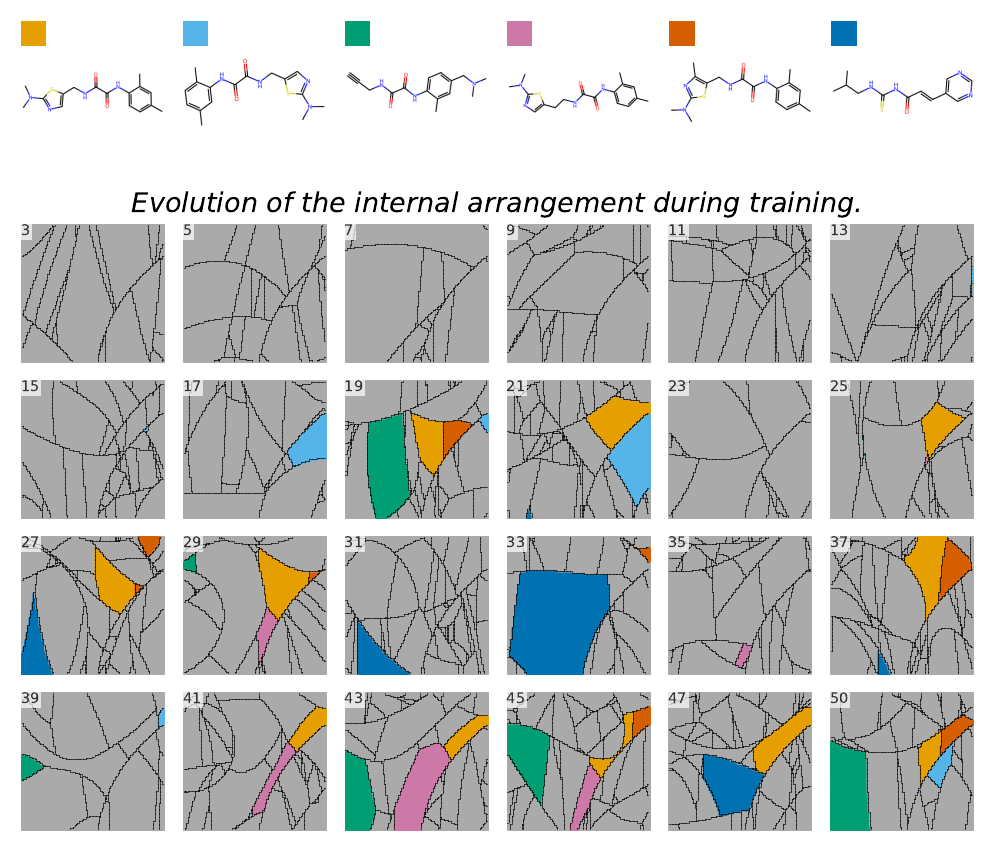}
    \caption{
    \textbf{Evolution of a MolMiner cross-section during training.}
    The same fixed two-dimensional section is decoded across different checkpoints. Regions are colored by molecular identity. The six most persistent identities across all snapshots are highlighted, while all other identities are shown in gray. An illustrative view of the evolution in Figure~\ref{fig:evolution-convergence}. Panel labels give the training epoch.
    }
    \label{fig:evolution-molminer}
\end{figure}

The same training dynamics can be inspected visually in fixed two-dimensional sections. Figures~\ref{fig:evolution-molminer} and~\ref{fig:evolution-hiervae} decode the same coordinate grid with the same random seed at every checkpoint. Early sections contain relatively few molecular identities and undergo substantial rearrangement. As training proceeds, MolMiner's identity count increases gradually and begins to level off, whereas HierVAE's rises rapidly before partially contracting. These trends mirror the neighborhood results.

The fixed sections are illustrative: each is a single two-dimensional view of a much higher-dimensional generative space. Within these sections, molecular regions shift, split, merge, expand, and contract during training. The highlighted persistent identities make some of this consolidation visible. Apparent connectedness or disappearance in a two-dimensional section should not, however, be interpreted as a statement about connectivity in the full space.

\section{Discussion}

Molecular generative models are usually evaluated through their outputs, while their internal coordinates are used for interpolation and optimization without first establishing how molecular identities are arranged within them. We address this missing step directly. Across three architectures, labeling decoded outputs by molecular identity reveals piecewise-constant regions in one- and two-dimensional cross-sections. These visualizations make explicit that a generative coordinate space is not simply a smooth field of molecules, but a partition whose boundaries determine when a change in coordinate becomes a change in molecular identity.

The existence of such regions is only the starting point. Before an internal space can be treated as a navigable chemical space, its organization must be evaluated with respect to the representation being probed, the equivalence convention used to define molecular identity, and the metric used to compare coordinates. Our results show why each choice matters. MolMiner and HierVAE retain coordinate-dependent chemical organization under stochastic decoding, whereas GDSS, under its standard stochastic sampler, places much of its exact-molecule variation in decoder randomness and shows clearer organization only under coarser scaffold or element conventions. HierVAE further shows that chemically cohesive neighborhoods need not be ordered by Euclidean distance or cosine similarity. The presence of an internal continuous coordinate space therefore does not, by itself, justify interpolation or local search. Navigation requires evidence that the chosen representation, identity resolution, and metric are aligned with the chemical question of interest. Our conclusions apply to the coordinates probed here, not to every hidden representation in each model; other internal spaces may organize molecular identity differently.

The explicit partitions also reveal structure that neighborhood statistics alone would miss. Across all three models, broad territories are subdivided by finer boundaries, suggesting a nested organization. In MolMiner, this pattern can be traced to the decoding process: early partial structures define broad lineages, later decisions subdivide them. The similar appearance of HierVAE and GDSS is consistent with an analogous divergence of shared trajectories. This branching picture provides a possible way to think about the formation of novel identities from shared partial trajectories.

Finally, training shows that chemical organization and local molecular granularity are distinct properties of the learned organization. In both MolMiner and HierVAE, cohesiveness stabilizes while the number of identities represented within a neighborhood continues to change. The model can therefore establish broad local chemical organization before its exact molecular repertoire has settled. These quantities offer complementary views of optimization: one measures how chemically localized a neighborhood is, while the other measures how finely it is subdivided.

\section{Conclusion}
\label{sec:conclusion}
We made molecular identity explicit and used it to examine how three molecular generative models organize the molecules they can produce. Across MolMiner, HierVAE, and GDSS, fixed-randomness probes reveal piecewise-constant molecular regions separated by sharp boundaries. The resulting picture is that internal organization is not a single property of a generative space. It depends on the representation being probed, the equivalence convention used to define molecular identity, and the metric used to navigate the coordinates. The coarse-to-fine boundaries observed across models also suggest that final molecular regions inherit structure from the decoding process. In MolMiner, this can be traced directly to the divergence of shared partial molecular trajectories. During training, chemical cohesiveness and molecular granularity evolve on different timescales, showing that broad chemical organization can stabilize while the exact molecular repertoire continues to change. These results provide a basis for measuring internal organization before assuming that a generative coordinate space supports chemically meaningful interpolation, local search, or optimization.

Beyond molecules, the framework applies wherever a generative model maps onto an output space equipped with a meaningful equivalence relation. Proteins are a natural next case~\cite{Ziegler2023}, and for crystals, continuous invariants under rigid motion provide candidate equivalence relations~\cite{WiddowsonK22,WiddowsonK21,ANOSOVA2026112108,Anosova_2024}.

More generally, this analysis provides a direct way to determine whether a generative space is organized, what its organization is defined with respect to, and whether it can be meaningfully navigated.

\section*{Data \& Code availability}
For the official implementations of the models used in this work, we refer to the original repositories: GDSS~\cite{pmlr-v162-jo22a}\footnote{\url{https://github.com/harryjo97/GDSS}}, HierVAE~\cite{jin2020hierarchicalgenerationmoleculargraphs}\footnote{\url{https://github.com/wengong-jin/hgraph2graph}}, and MolMiner~\cite{ortegaochoa2025molminercontrollable3dawarefragmentbased}\footnote{\url{https://github.com/raulorteg/molminer}}. The analysis scripts used in this paper will be released at \url{https://github.com/raulorteg/molecular-identity-partitions} upon publication.

\section*{Computational requirements}
HierVAE and MolMiner were retrained in-house on a single NVIDIA RTX 3090 GPU. All latent space sampling, post-processing, and data analysis were performed on a standard laptop (Apple MacBook Pro, Apple M4 Max).

\section*{Competing interests}
The authors declare no competing interests.

\section*{Acknowledgments}
The authors acknowledge support from the Pioneer Center for Accelerating P2X Materials Discovery (CAPeX), DNRF grant number P3, the Novo Nordisk Foundation Grant no. NNF24OC0089800 and NNF22OC0078009. We thank Steven B. Torrisi for helpful discussions.

\bibliographystyle{unsrtnat}
\bibliography{bibliography}
\newpage

\appendix
\renewcommand{\thefigure}{S\arabic{figure}}
\setcounter{figure}{0}
\renewcommand{\thetable}{S\arabic{table}}
\setcounter{table}{0}
\renewcommand{\theequation}{S\arabic{equation}}
\setcounter{equation}{0}

\begin{center}
{\Large\textbf{Supporting Information}}\\[1ex]
{\large How Molecular Generative Models Organize Molecular Identity}\\[1ex]
\end{center}

\section{Robustness of identity-cell structure to slice direction}\label{sec:si-alt-slices}
\begin{figure}[htbp]
    \centering
    \begin{subfigure}[t]{0.19\textwidth}
        \centering
        \includegraphics[width=\textwidth]{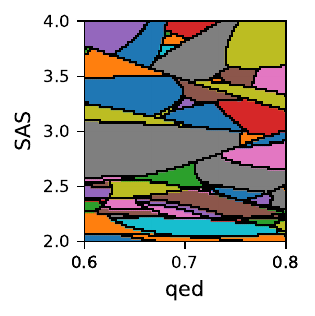}
        \caption{qed $\times$ SAS.}
    \end{subfigure}
    \begin{subfigure}[t]{0.19\textwidth}
        \centering
        \includegraphics[width=\textwidth]{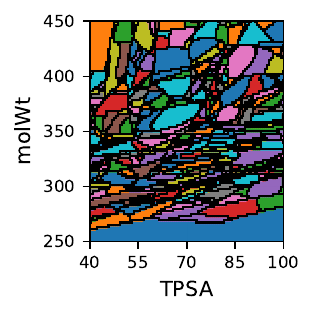}
        \caption{TPSA $\times$ molWt.}
    \end{subfigure}
    \begin{subfigure}[t]{0.19\textwidth}
        \centering
        \includegraphics[width=\textwidth]{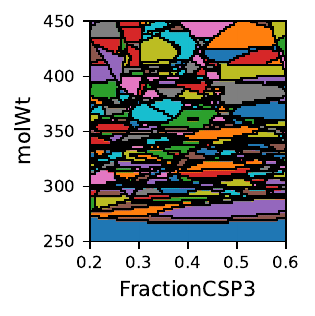}
        \caption{FractionCSP3 $\times$ molWt.}
    \end{subfigure}
    \begin{subfigure}[t]{0.19\textwidth}
        \centering
        \includegraphics[width=\textwidth]{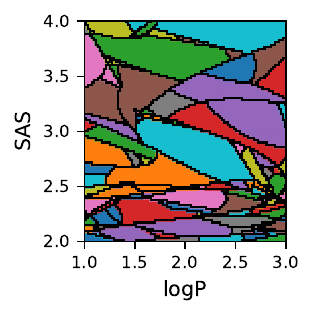}
        \caption{logP $\times$ SAS.}
    \end{subfigure}
    \begin{subfigure}[t]{0.19\textwidth}
        \centering
        \includegraphics[width=\textwidth]{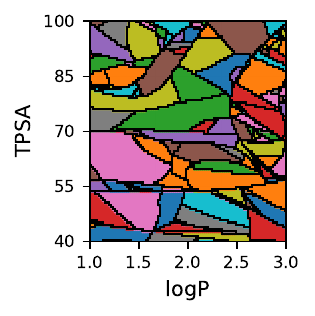}
        \caption{logP $\times$ TPSA.}
    \end{subfigure}
    \caption{\textbf{MolMiner 2D slices across alternative property-pair axes.} Same anchor as Fig.~\ref{fig:deterministic-tessellation}(a), with $\approx 10\,000$ samples and each $100\times100$ slice taken over a pair of channels of the $12$-dim conditioning vector.}
    \label{fig:si-molminer-slices}
\end{figure}

\begin{figure}[htbp]
      \centering
      \begin{subfigure}[t]{0.19\textwidth}
          \centering
          \includegraphics[width=\textwidth]{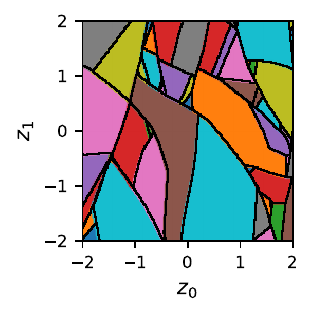}
          \caption{$z_0 \times z_1$.}
          \label{fig:hv_sub01}
      \end{subfigure}
      \begin{subfigure}[t]{0.19\textwidth}
          \centering
          \includegraphics[width=\textwidth]{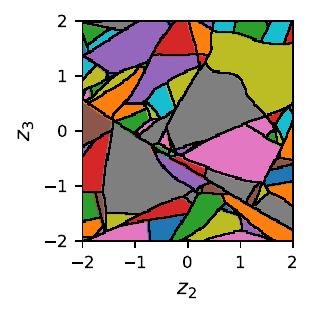}
          \caption{$z_2 \times z_3$.}
          \label{fig:hv_sub23}
      \end{subfigure}
      \hfill
      \begin{subfigure}[t]{0.19\textwidth}
          \centering
          \includegraphics[width=\textwidth]{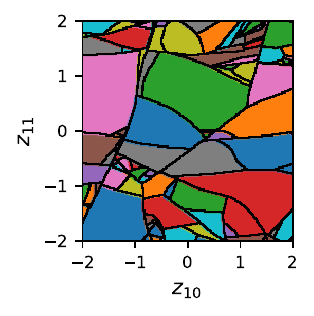}
          \caption{$z_{10} \times z_{11}$.}
          \label{fig:hv_sub1011}
      \end{subfigure}
      \hfill
      \begin{subfigure}[t]{0.19\textwidth}
          \centering
          \includegraphics[width=\textwidth]{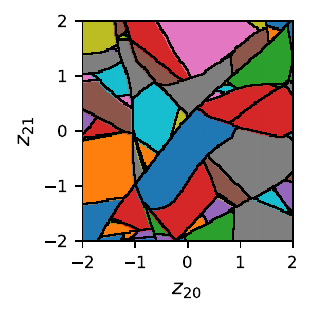}
          \caption{$z_{20} \times z_{21}$.}
          \label{fig:hv_sub2021}
      \end{subfigure}
      \hfill
      \begin{subfigure}[t]{0.19\textwidth}
          \centering
          \includegraphics[width=\textwidth]{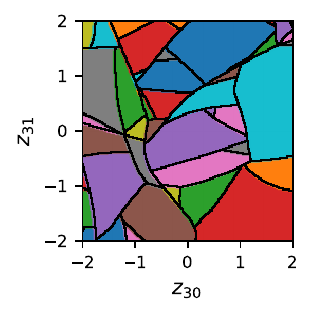}
          \caption{$z_{30} \times z_{31}$.}
          \label{fig:hv_sub3031}
      \end{subfigure}
      \caption{\textbf{HierVAE 2D slices across alternative latent-dim pairs.} Same anchor as Fig.~\ref{fig:deterministic-tessellation}(b), $\approx 40\,000$ samples ($200\times200$ slice), remaining latent dimensions fixed, with each slice perturbing a different pair of dimensions of the VAE latent.}
      \label{fig:hv_si}
\end{figure}

\begin{figure}[htbp]
      \centering
      \begin{subfigure}[t]{0.19\textwidth}
          \centering
          \includegraphics[width=\textwidth]{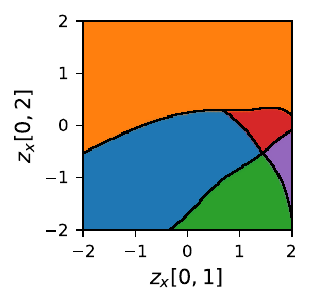}
          \caption{$z_x[0,1] \times z_x[0,2]$.}
          \label{fig:gdss_sub_xb}
      \end{subfigure}
      \hfill
      \begin{subfigure}[t]{0.19\textwidth}
          \centering
          \includegraphics[width=\textwidth]{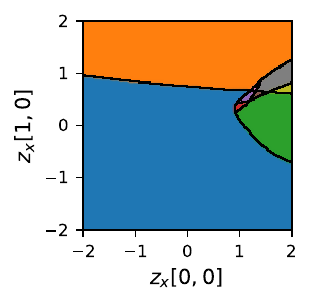}
          \caption{$z_x[0,0] \times z_x[1,0]$.}
          \label{fig:gdss_sub_xc}
      \end{subfigure}
      \hfill
      \begin{subfigure}[t]{0.19\textwidth}
          \centering
          \includegraphics[width=\textwidth]{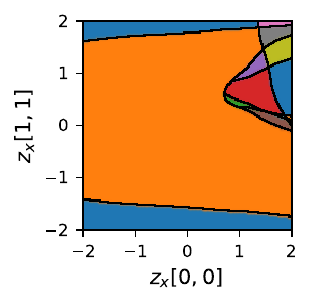}
          \caption{$z_x[0,0] \times z_x[1,1]$.}
          \label{fig:gdss_sub_xd}
      \end{subfigure}
      \begin{subfigure}[t]{0.19\textwidth}
          \centering
          \includegraphics[width=\textwidth]{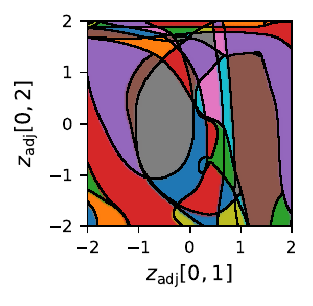}
          \caption{$z_{\mathrm{adj}}[0,1] \times z_{\mathrm{adj}}[0,2]$.}
          \label{fig:gdss_sub_adja}
      \end{subfigure}
      \hfill
      \begin{subfigure}[t]{0.19\textwidth}
          \centering
          \includegraphics[width=\textwidth]{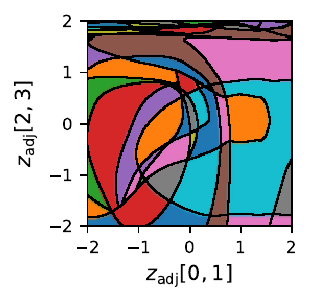}
          \caption{$z_{\mathrm{adj}}[0,1] \times z_{\mathrm{adj}}[2,3]$.}
          \label{fig:gdss_sub_adjb}
      \end{subfigure}

      \caption{\textbf{GDSS 2D slices across alternative axes of the prior-noise.} (24 active nodes). $\approx 40\,000$ samples ($200\times200$ slice). Each slice perturbs a different pair in $z_x$ (one-hot atom-feature channels; panels a--c) or $z_{\mathrm{adj}}$ (adjacency channel; panels d--e). Atom-type $z_x$ perturbations result in coarser partitions (1--13 cells per slice) than perturbations of the adjacency axis (61--65 cells per slice), illustrating that the partition is slice-direction dependent.}
      \label{fig:gdss_si}
  \end{figure}

The 2D slices in Figure~\ref{fig:deterministic-tessellation}(a--c) are taken along a single fixed pair of coordinates per model. To verify that the partition is not specific to that choice, we repeat it along alternative pairs of axes in each model. Across all three models, piecewise-constant structure persists across slice directions; what changes is the density of the tiling and its orientation. For Figs.~\ref{fig:si-molminer-slices}--\ref{fig:gdss_si} note that colors cycle within each panel and do not identify the same molecule across panels.

\newpage
\section{Alternative equivalence relations}\label{sec:si-equiv-relations}

\begin{figure}[htbp]
    \centering
    \begin{subfigure}[t]{0.16\textwidth}
        \centering
        \includegraphics[width=\textwidth]{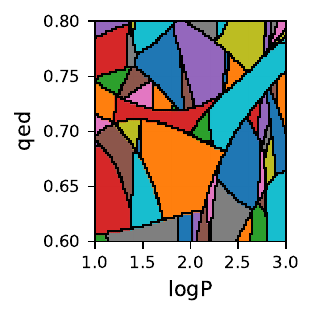}
        \caption{SMILES\\ (\(60\) identities)}
        \label{fig:molminer-equiv-canonical}
    \end{subfigure}
    \hfill
    \begin{subfigure}[t]{0.16\textwidth}
        \centering
        \includegraphics[width=\textwidth]{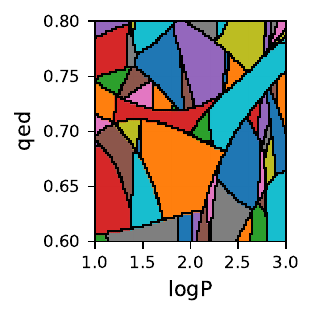}
        \caption{InChIKey-14\\ (\(60\) identities)}
        \label{fig:molminer-equiv-inchikey}
    \end{subfigure}
    \hfill
    \begin{subfigure}[t]{0.16\textwidth}
        \centering
        \includegraphics[width=\textwidth]{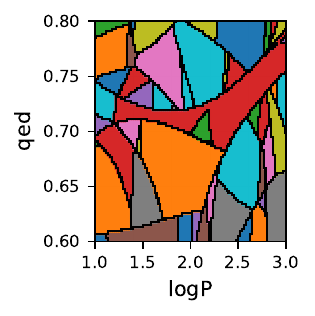}
        \caption{Formula\\ (\(45\) identities)}
        \label{fig:molminer-equiv-formula}
    \end{subfigure}
    \hfill
    \begin{subfigure}[t]{0.16\textwidth}
        \centering
        \includegraphics[width=\textwidth]{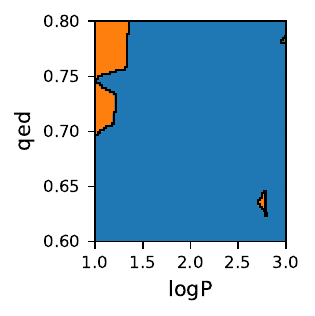}
        \caption{Elements\\ (\(3\) identities)}
        \label{fig:molminer-equiv-composition}
    \end{subfigure}
    \hfill
    \begin{subfigure}[t]{0.16\textwidth}
        \centering
        \includegraphics[width=\textwidth]{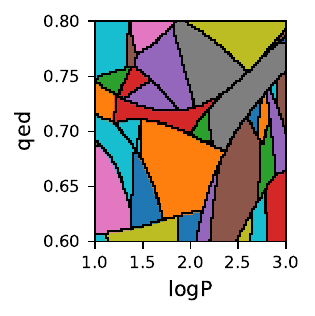}
        \caption{Murcko\\ (\(31\) identities)}
        \label{fig:molminer-equiv-murcko}
    \end{subfigure}
    \hfill
    \begin{subfigure}[t]{0.16\textwidth}
        \centering
        \includegraphics[width=\textwidth]{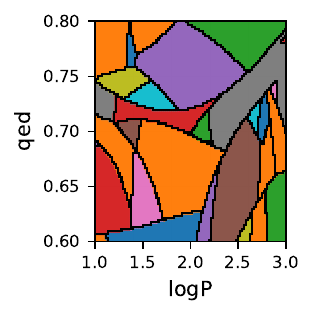}
        \caption{Generic Murcko\\ (\(24\) identities)}
        \label{fig:molminer-equiv-murcko-generic}
    \end{subfigure}

    \caption{
    \textbf{Changing the molecular equivalence relation coarsens the MolMiner partition.}
    The same fixed MolMiner two-dimensional slice is decoded once and labeled under six equivalence conventions. Each panel shows the resulting partition.
    }
    \label{fig:molminer-equiv-quotients}
\end{figure}

\begin{figure}[htbp]
    \centering
    \begin{subfigure}[t]{0.16\textwidth}
        \centering
        \includegraphics[width=\textwidth]{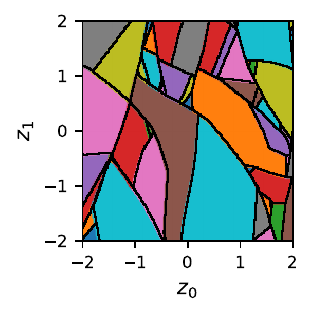}
        \caption{SMILES\\ (\(61\) identities)}
        \label{fig:hiervae-equiv-canonical}
    \end{subfigure}
    \hfill
    \begin{subfigure}[t]{0.16\textwidth}
        \centering
        \includegraphics[width=\textwidth]{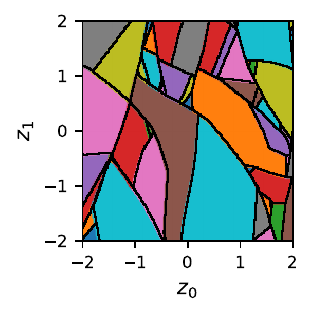}
        \caption{InChIKey-14\\ (\(61\) identities)}
        \label{fig:hiervae-equiv-inchikey}
    \end{subfigure}
    \hfill
    \begin{subfigure}[t]{0.16\textwidth}
        \centering
        \includegraphics[width=\textwidth]{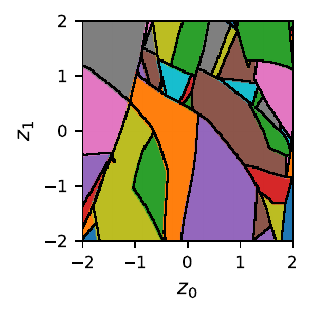}
        \caption{Formula\\ (\(44\) identities)}
        \label{fig:hiervae-equiv-formula}
    \end{subfigure}
    \hfill
    \begin{subfigure}[t]{0.16\textwidth}
        \centering
        \includegraphics[width=\textwidth]{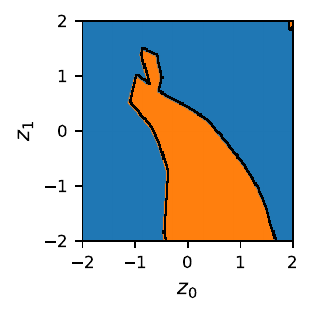}
        \caption{Elements\\ (\(2\) identities)}
        \label{fig:hiervae-equiv-composition}
    \end{subfigure}
    \hfill
    \begin{subfigure}[t]{0.16\textwidth}
        \centering
        \includegraphics[width=\textwidth]{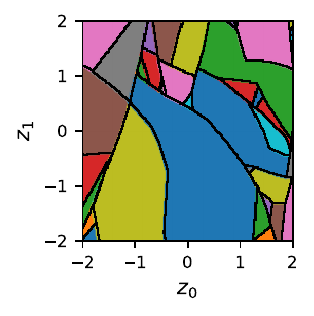}
        \caption{Murcko\\ (\(41\) identities)}
        \label{fig:hiervae-equiv-murcko}
    \end{subfigure}
    \hfill
    \begin{subfigure}[t]{0.16\textwidth}
        \centering
        \includegraphics[width=\textwidth]{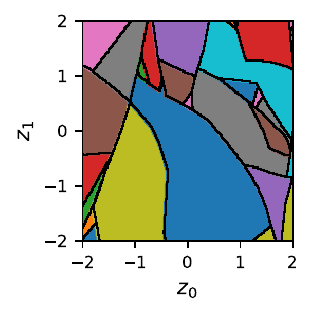}
        \caption{Generic Murcko\\ (\(31\) identities)}
        \label{fig:hiervae-equiv-murcko-generic}
    \end{subfigure}

    \caption{
    \textbf{Changing the molecular equivalence relation coarsens the HierVAE partition.}
    The same fixed HierVAE two-dimensional slice is decoded once and labeled under six output equivalence conventions. Each panel shows the resulting partition.
    }
    \label{fig:hiervae-equiv-quotients}
\end{figure}

\begin{figure}[htbp]
    \centering
    \begin{subfigure}[t]{0.16\textwidth}
        \centering
        \includegraphics[width=\textwidth]{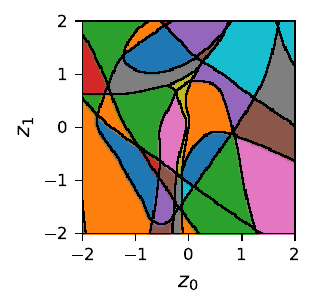}
        \caption{SMILES\\ (\(44\) identities)}
        \label{fig:gdss-equiv-canonical}
    \end{subfigure}
    \hfill
    \begin{subfigure}[t]{0.16\textwidth}
        \centering
        \includegraphics[width=\textwidth]{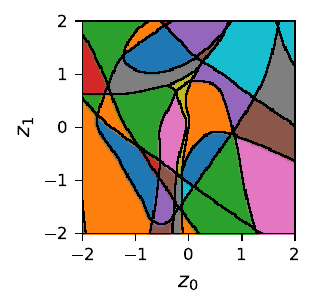}
        \caption{InChIKey-14\\ (\(44\) identities)}
        \label{fig:gdss-equiv-inchikey}
    \end{subfigure}
    \hfill
    \begin{subfigure}[t]{0.16\textwidth}
        \centering
        \includegraphics[width=\textwidth]{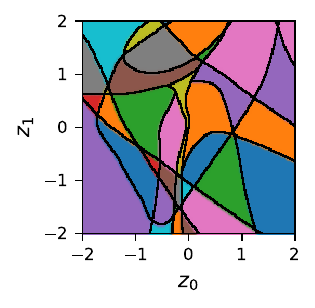}
        \caption{Formula\\ (\(37\) identities)}
        \label{fig:gdss-equiv-formula}
    \end{subfigure}
    \hfill
    \begin{subfigure}[t]{0.16\textwidth}
        \centering
        \includegraphics[width=\textwidth]{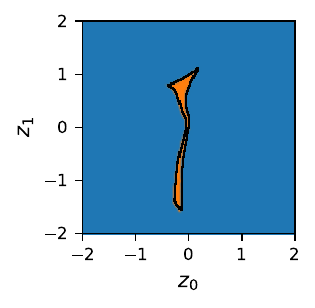}
        \caption{Elements\\ (\(2\) identities)}
        \label{fig:gdss-equiv-composition}
    \end{subfigure}
    \hfill
    \begin{subfigure}[t]{0.16\textwidth}
        \centering
        \includegraphics[width=\textwidth]{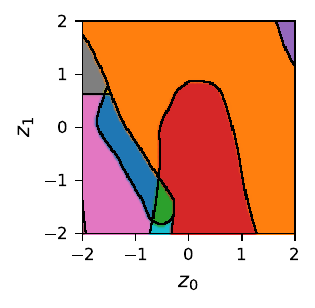}
        \caption{Murcko\\ (\(10\) identities)}
        \label{fig:gdss-equiv-murcko}
    \end{subfigure}
    \hfill
    \begin{subfigure}[t]{0.16\textwidth}
        \centering
        \includegraphics[width=\textwidth]{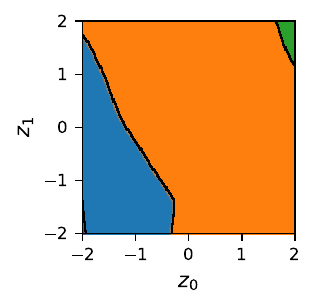}
        \caption{Generic Murcko\\ (\(4\) identities)}
        \label{fig:gdss-equiv-murcko-generic}
    \end{subfigure}

    \caption{
    \textbf{Changing the molecular equivalence relation coarsens the GDSS partition.}
    The same fixed GDSS two-dimensional prior-noise slice is decoded once and re-labeled under six equivalence conventions. Each panel shows the resulting partition. The strong collapse from SMILES identity to Murcko and generic Murcko indicates that much of the apparent exact-identity variation in this fixed-\(\eta\) GDSS section shares a smaller number of scaffold or scaffold-topology classes.
    }
    \label{fig:gdss-equiv-quotients}
\end{figure}

\begin{figure}[htbp]
    \centering
    \begin{subfigure}[t]{0.48\textwidth}
        \centering
        \includegraphics[width=\textwidth]{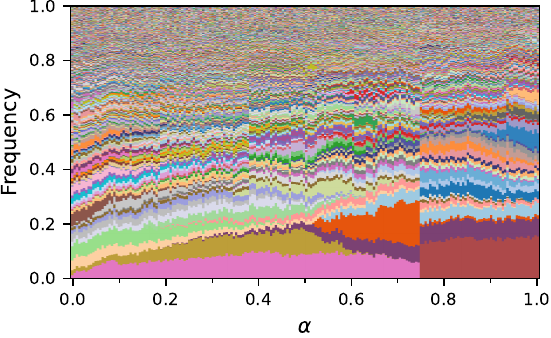}
        \caption{SMILES}
        \label{fig:flow-molminer-canonical}
    \end{subfigure}
    \hfill
    \begin{subfigure}[t]{0.48\textwidth}
        \centering
        \includegraphics[width=\textwidth]{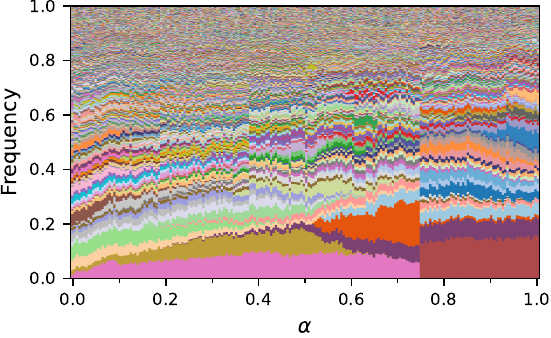}
        \caption{InChIKey-14}
        \label{fig:flow-molminer-inchikey}
    \end{subfigure}
    \vspace{0.5em}
    \begin{subfigure}[t]{0.48\textwidth}
        \centering
        \includegraphics[width=\textwidth]{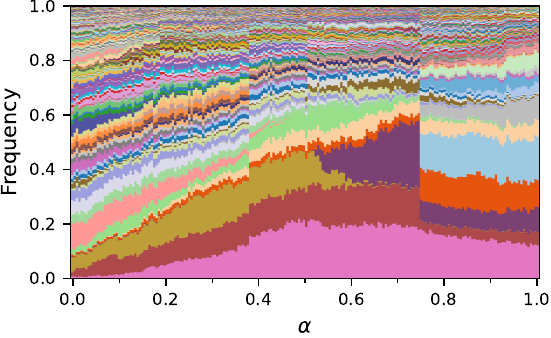}
        \caption{Formula}
        \label{fig:flow-molminer-formula}
    \end{subfigure}
    \hfill
    \begin{subfigure}[t]{0.48\textwidth}
        \centering
        \includegraphics[width=\textwidth]{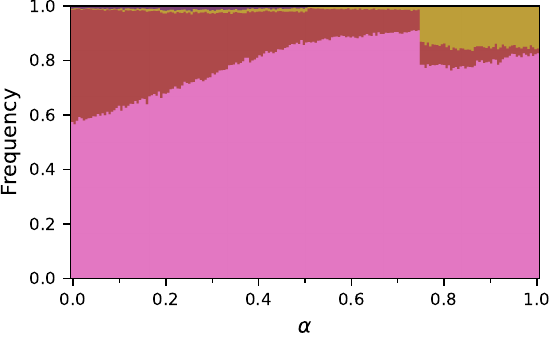}
        \caption{Elements}
        \label{fig:flow-molminer-composition}
    \end{subfigure}
    \vspace{0.5em}
    \begin{subfigure}[t]{0.48\textwidth}
        \centering
        \includegraphics[width=\textwidth]{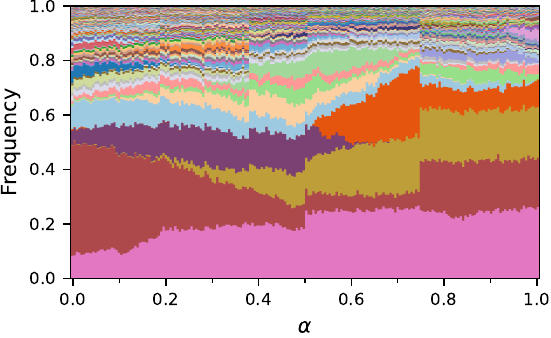}
        \caption{Murcko}
        \label{fig:flow-molminer-murcko}
    \end{subfigure}
    \hfill
    \begin{subfigure}[t]{0.48\textwidth}
        \centering
        \includegraphics[width=\textwidth]{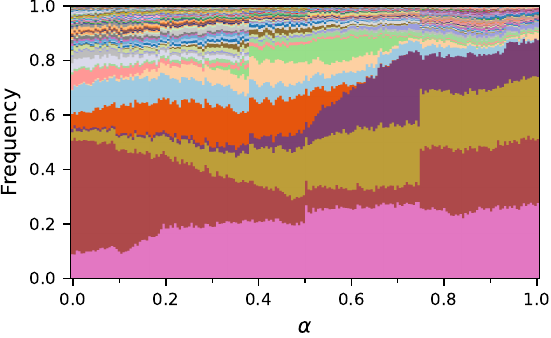}
        \caption{Generic Murcko}
        \label{fig:flow-molminer-murcko-generic}
    \end{subfigure}
    \caption{
    \textbf{Dependence of decoded molecular identities along a path in MolMiner under alternative equivalence conventions (stochastic).} The straight-line path in \(\mathcal{Z}\) shown in Fig.~\ref{fig:prob-decoding-molminer} is labeled under the six molecular-equivalence conventions. Segment heights reflect relative class frequencies. Colors identify classes within each panel only.
    }
    \label{fig:si-flow-molminer-equiv}
\end{figure}

\begin{figure}[htbp]
    \centering
    \begin{subfigure}[t]{0.48\textwidth}
        \centering
        \includegraphics[width=\textwidth]{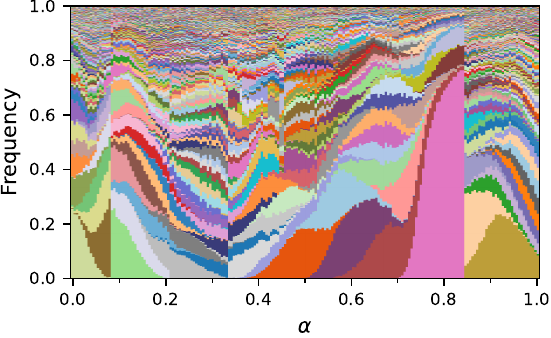}
        \caption{SMILES}
        \label{fig:flow-hiervae-canonical}
    \end{subfigure}
    \hfill
    \begin{subfigure}[t]{0.48\textwidth}
        \centering
        \includegraphics[width=\textwidth]{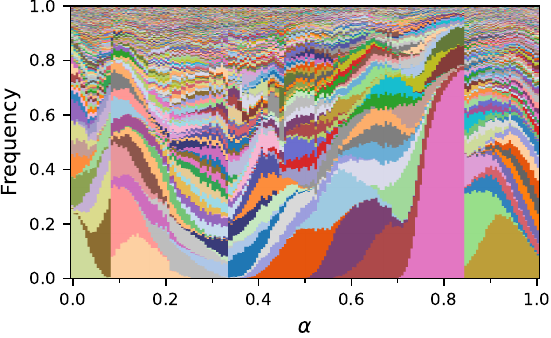}
        \caption{InChIKey-14}
        \label{fig:flow-hiervae-inchikey}
    \end{subfigure}

    \vspace{0.5em}

    \begin{subfigure}[t]{0.48\textwidth}
        \centering
        \includegraphics[width=\textwidth]{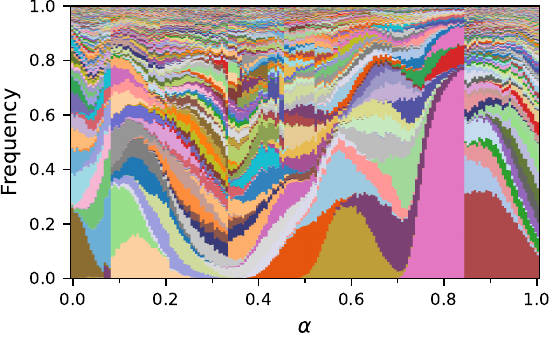}
        \caption{Formula}
        \label{fig:flow-hiervae-formula}
    \end{subfigure}
    \hfill
    \begin{subfigure}[t]{0.48\textwidth}
        \centering
        \includegraphics[width=\textwidth]{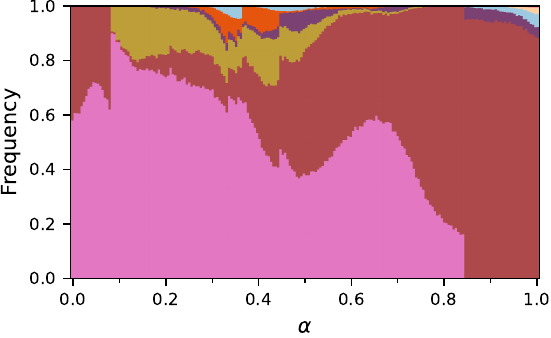}
        \caption{Elements}
        \label{fig:flow-hiervae-composition}
    \end{subfigure}

    \vspace{0.5em}

    \begin{subfigure}[t]{0.48\textwidth}
        \centering
        \includegraphics[width=\textwidth]{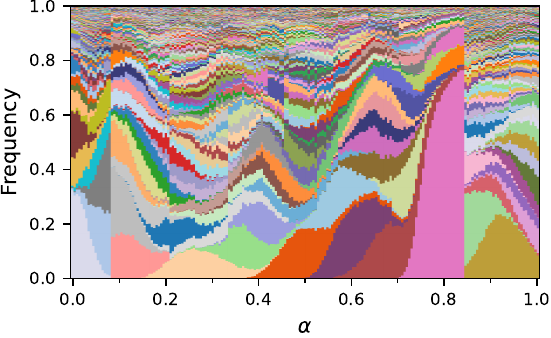}
        \caption{Murcko}
        \label{fig:flow-hiervae-murcko}
    \end{subfigure}
    \hfill
    \begin{subfigure}[t]{0.48\textwidth}
        \centering
        \includegraphics[width=\textwidth]{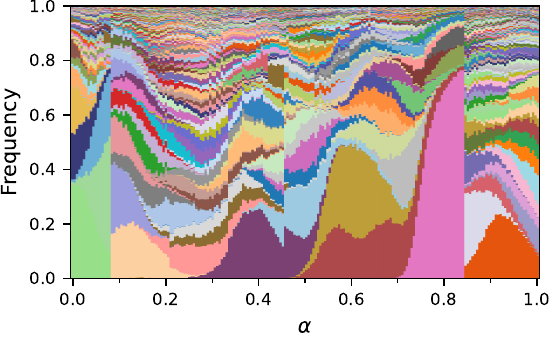}
        \caption{Generic Murcko}
        \label{fig:flow-hiervae-murcko-generic}
    \end{subfigure}

    \caption{
    \textbf{Dependence of decoded molecular identities along a path in HierVAE under alternative equivalence conventions (stochastic).}
    The straight-line path in \(\mathcal{Z}\) is labeled under the six molecular-equivalence conventions. Segment heights reflect relative class frequencies. Colors identify classes within each panel only.
    }
    \label{fig:si-flow-hiervae-equiv}
\end{figure}

\begin{figure}[htbp]
    \centering
    \begin{subfigure}[t]{0.48\textwidth}
        \centering
        \includegraphics[width=\textwidth]{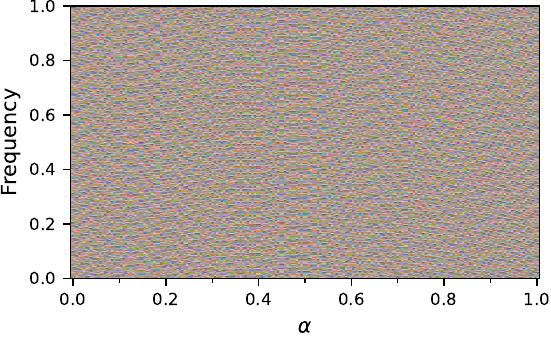}
        \caption{SMILES}
        \label{fig:flow-gdss-canonical}
    \end{subfigure}
    \hfill
    \begin{subfigure}[t]{0.48\textwidth}
        \centering
        \includegraphics[width=\textwidth]{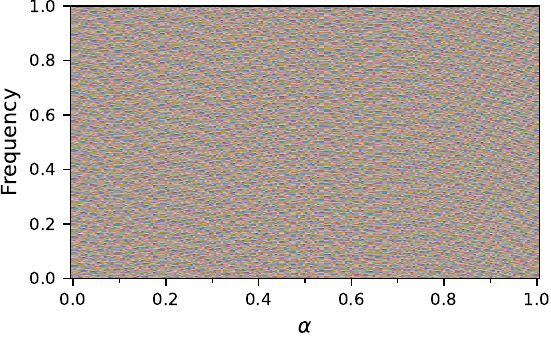}
        \caption{InChIKey-14}
        \label{fig:flow-gdss-inchikey}
    \end{subfigure}
    \vspace{0.5em}
    \begin{subfigure}[t]{0.48\textwidth}
        \centering
        \includegraphics[width=\textwidth]{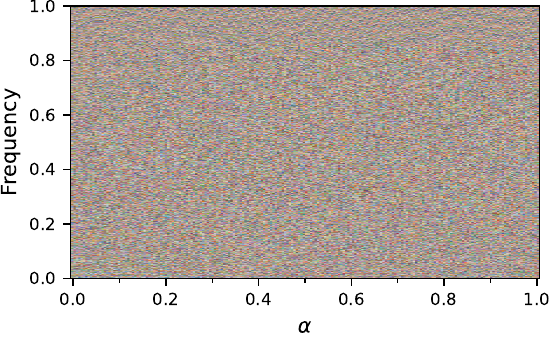}
        \caption{Formula}
        \label{fig:flow-gdss-formula}
    \end{subfigure}
    \hfill
    \begin{subfigure}[t]{0.48\textwidth}
        \centering
        \includegraphics[width=\textwidth]{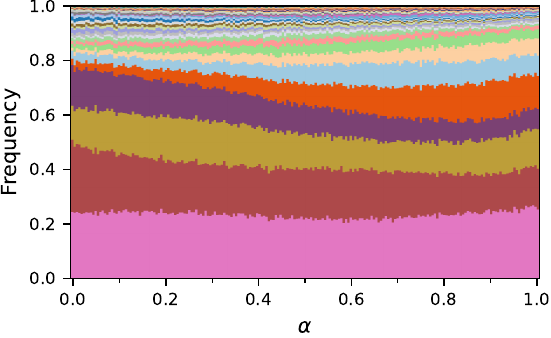}
        \caption{Elements}
        \label{fig:flow-gdss-composition}
    \end{subfigure}
    \vspace{0.5em}
    \begin{subfigure}[t]{0.48\textwidth}
        \centering
        \includegraphics[width=\textwidth]{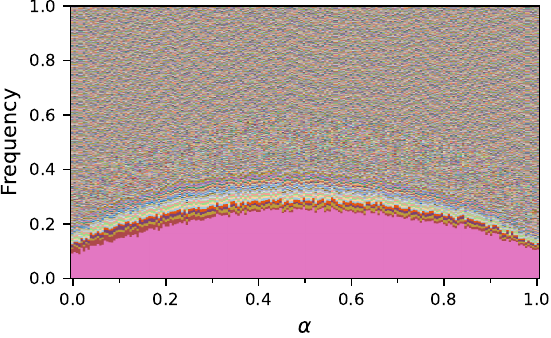}
        \caption{Murcko}
        \label{fig:flow-gdss-murcko}
    \end{subfigure}
    \hfill
    \begin{subfigure}[t]{0.48\textwidth}
        \centering
        \includegraphics[width=\textwidth]{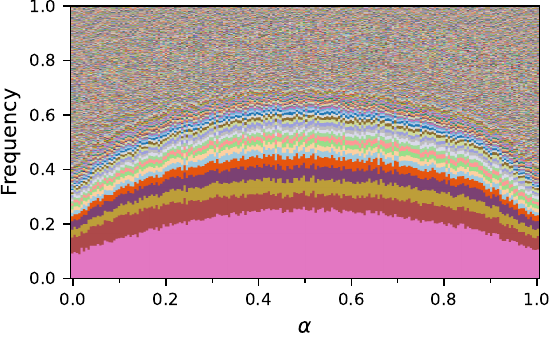}
        \caption{Generic Murcko}
        \label{fig:flow-gdss-murcko-generic}
    \end{subfigure}
    \caption{
    \textbf{Dependence of decoded molecular identities along a path in GDSS under alternative equivalence conventions (stochastic).}
    Segment heights reflect the relative frequency of each equivalence class at each path point. Colors are not comparable across panels.
    }
    \label{fig:si-flow-gdss-equiv}
\end{figure}

We re-labeled the decoded molecules produced in Figures~\ref{fig:deterministic-tessellation},~\ref{fig:prob-decoding} under all six equivalence conventions, showing the same partition at different resolutions. First, we re-label fixed two-dimensional sections in Figures~\ref{fig:molminer-equiv-quotients}, \ref{fig:hiervae-equiv-quotients}, and~\ref{fig:gdss-equiv-quotients}. Then, we re-label Figure~\ref{fig:prob-decoding}, resulting in Figures~\ref{fig:si-flow-molminer-equiv}, \ref{fig:si-flow-hiervae-equiv}, and~\ref{fig:si-flow-gdss-equiv}. 

Across all three architectures, coarsening the relation merges territories while preserving the broad path-dependence of the partition. GDSS is the exception: at SMILES and InChIKey-14 resolution its flows are effectively structureless, repeated decoding yields near-unique identities. Organization within GDSS emerges only once the relation is coarsened to scaffold or element, and even then the path-dependence is comparatively weaker than MolMiner and HierVAE's. Colors encode identities within each panel only and are not comparable across panels.

\section{Example of order-agnostic trajectories in MolMiner}\label{sec:si-order-agnostic}
\begin{figure}[H]
    \centering
    \includegraphics[width=\linewidth]{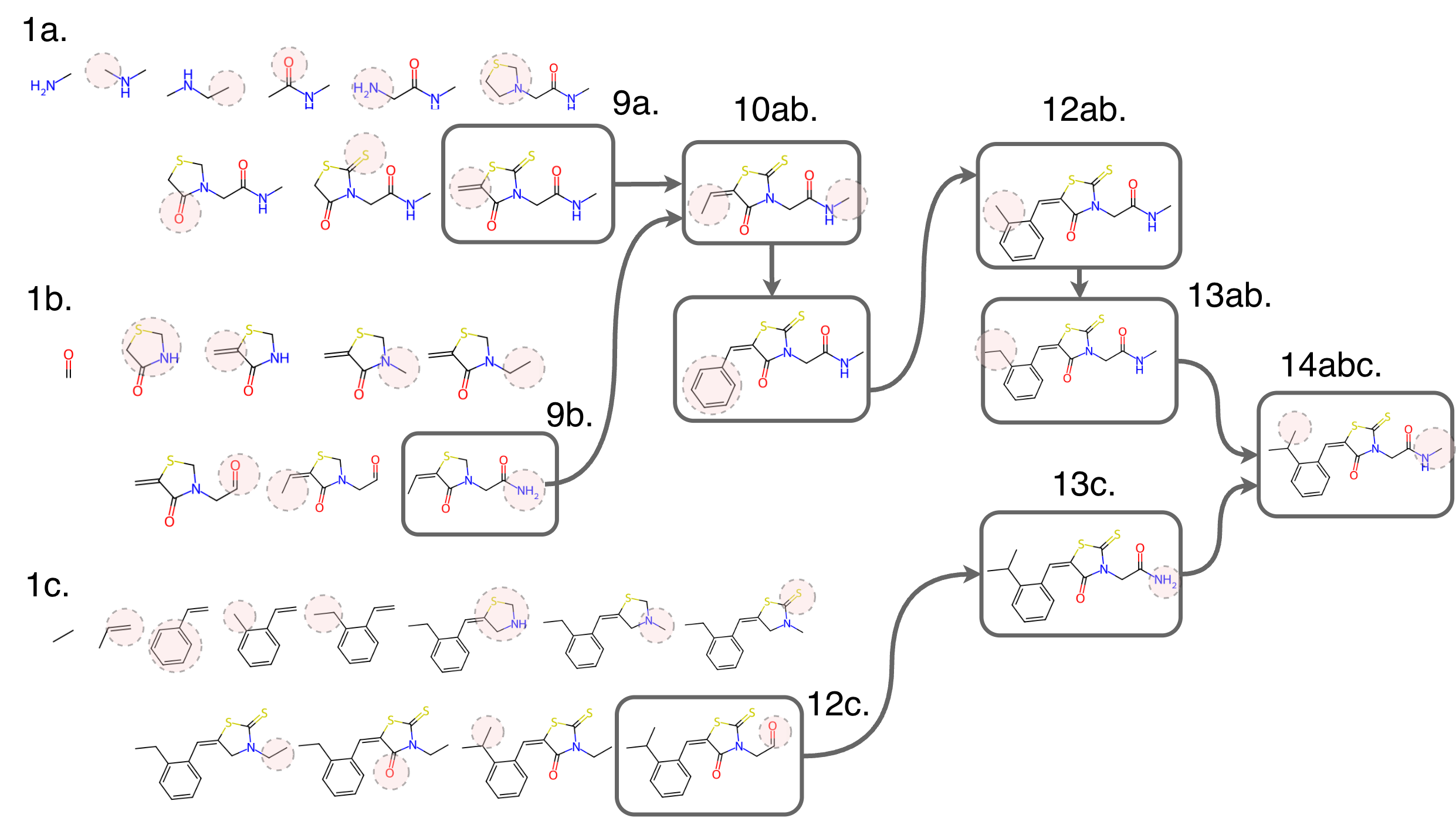}
    \caption{\textbf{Order-agnostic generation in MolMiner.} A single coordinate \(z\in\mathcal Z\) was fixed and decoded repeatedly; some molecules were reached through more than one trajectory. We show one representative example. Its three trajectories (1a--1c) reach the same canonical-SMILES through different generation orders. Dashed circles mark the fragment added at each step.
}
    \label{fig:si-traj}
\end{figure}

\section{Additional training-time evolution of the partition}\label{si:tesstrain-hiervae}
\begin{figure}[H]
    \centering
    \includegraphics[width=0.9\linewidth]{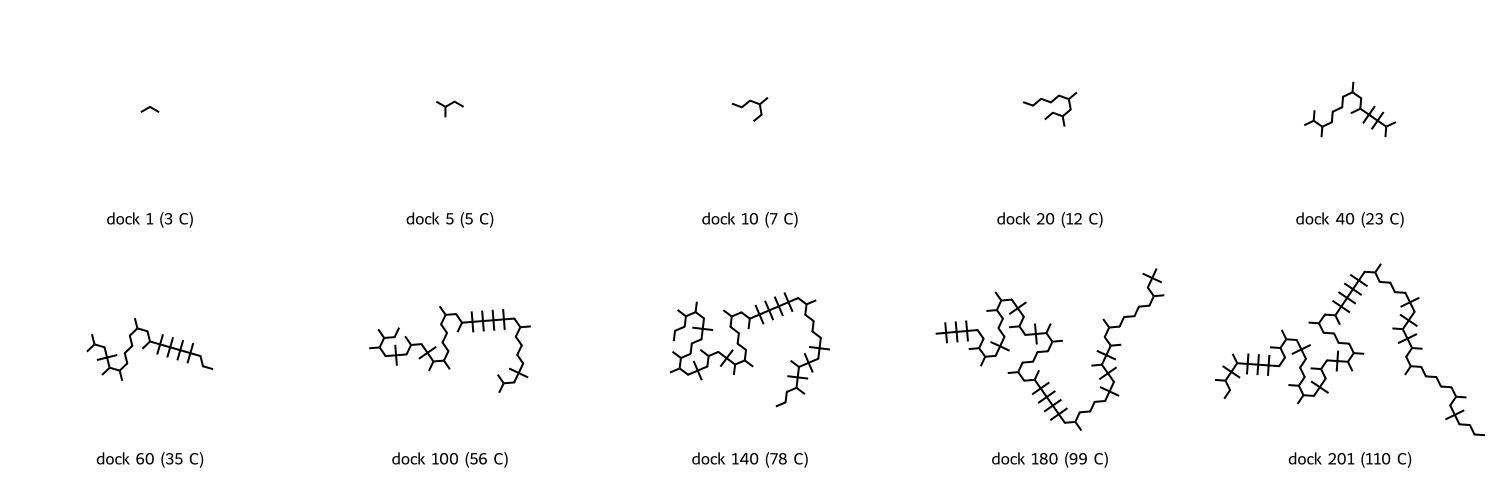}
    \caption{\textbf{Decode runaway}. Lacking a learned termination signal (epoch 1), the decoder appends fragments indefinitely growing a $C_{110}$ alkane before failing.}
    \label{fig:runaway}
\end{figure}

Tables~\ref{tab:molminer_ball_convergence} and~\ref{tab:hiervae_ball_convergence} give the full per-checkpoint breakdown of the neighborhood analysis summarized in Section~\ref{subsec:results-training} for MolMiner and HierVAE. SMILES and InChIKey-14 cross-neighborhood overlap is zero at every checkpoint in both models, and appreciable overlap appears only at the element level. Cohesiveness and granularity converge on different timescales. In MolMiner, \(\mathrm{AUC}(W,A)\) reaches its \(0.83\)--\(0.86\) plateau by roughly epoch 5, after which it is flat, while \(n_{\mathrm{unique}}\) continues to grow from \(\sim 200\) to \(\sim 430\). HierVAE shows the same decoupling with a different granularity trajectory: \(\mathrm{AUC}(W,A)\) rises more gradually to its \(0.87\)--\(0.88\) plateau (\(\sim\)50k steps), while \(n_{\mathrm{unique}}\) overshoots near 100k steps and then decreases, so the two quantities are not monotonically coupled. MolMiner statistics are computed on \(K=20\) neighborhoods that decode to completion at all checkpoints; Fig.~\ref{fig:runaway} shows one of the runaway molecules.
\begin{table}[H]
  \centering
  \footnotesize
  \caption{MolMiner partition organization across training. Left: cross-neighborhood median Jaccard per equivalence convention. Right: within- and across-neighborhood ECFP Tanimoto. Each cell reports the median$^{Q_3}_{Q_1}$ over neighborhoods.}
  \label{tab:molminer_ball_convergence}
  \resizebox{\linewidth}{!}{%
  \begin{tabular}{lcccccccccc}
    \toprule
    & \multicolumn{6}{c}{Cross-neighborhood Jaccard} & \multicolumn{4}{c}{ECFP cohesiveness} \\
    \cmidrule(lr){2-7} \cmidrule(lr){8-11}
    epoch & SMILES & InChIKey-14 & murcko & murcko-gen & formula & element & med$(W)$ & med$(A)$ & AUC$_{W,A}$ & $n_{\mathrm{unique}}$ \\
    \midrule
    $1$ & $0$ & $0$ & $0^{0.17}_{0}$ & $0^{0.17}_{0}$ & $0^{0.0089}_{0}$ & $0.33^{0.50}_{0}$ & $0.27^{0.39}_{0.22}$ & $0.17^{0.18}_{0.14}$ & $0.74$ & $9^{19}_{2}$ \\
    $2$ & $0$ & $0$ & $0.0063^{0.022}_{0}$ & $0.025^{0.048}_{0}$ & $0^{0.017}_{0}$ & $0.50^{0.67}_{0.25}$ & $0.28^{0.34}_{0.24}$ & $0.20^{0.20}_{0.19}$ & $0.80$ & $50^{85}_{27}$ \\
    $3$ & $0$ & $0$ & $0^{0.013}_{0}$ & $0.017^{0.048}_{0}$ & $0^{0.019}_{0}$ & $0.40^{0.67}_{0.25}$ & $0.24^{0.30}_{0.22}$ & $0.18^{0.18}_{0.16}$ & $0.80$ & $89^{134}_{48}$ \\
    $4$ & $0$ & $0$ & $0^{0.0097}_{0}$ & $0.017^{0.038}_{0}$ & $0^{0.0063}_{0}$ & $0.25^{0.40}_{0.14}$ & $0.23^{0.27}_{0.21}$ & $0.16^{0.17}_{0.15}$ & $0.83$ & $174^{239}_{47}$ \\
    $5$ & $0$ & $0$ & $0.0033^{0.0067}_{0}$ & $0.022^{0.044}_{0.0067}$ & $0^{0.012}_{0}$ & $0.33^{0.50}_{0.22}$ & $0.22^{0.24}_{0.21}$ & $0.14^{0.16}_{0.13}$ & $0.84$ & $204^{388}_{138}$ \\
    $10$ & $0$ & $0$ & $0.0044^{0.011}_{0}$ & $0.029^{0.052}_{0.0016}$ & $0^{0.0086}_{0}$ & $0.27^{0.40}_{0.17}$ & $0.22^{0.26}_{0.19}$ & $0.13^{0.14}_{0.12}$ & $0.86$ & $267^{397}_{100}$ \\
    $15$ & $0$ & $0$ & $0.0040^{0.0076}_{0}$ & $0.024^{0.052}_{0.011}$ & $0^{0.0076}_{0}$ & $0.29^{0.43}_{0.18}$ & $0.21^{0.24}_{0.19}$ & $0.13^{0.15}_{0.12}$ & $0.84$ & $258^{362}_{161}$ \\
    $20$ & $0$ & $0$ & $0.0037^{0.0079}_{0}$ & $0.029^{0.050}_{0.012}$ & $0^{0.0079}_{0}$ & $0.29^{0.40}_{0.17}$ & $0.21^{0.25}_{0.19}$ & $0.12^{0.14}_{0.11}$ & $0.84$ & $323^{455}_{217}$ \\
    $25$ & $0$ & $0$ & $0.0026^{0.0066}_{0}$ & $0.020^{0.042}_{0.0068}$ & $0^{0.0084}_{0}$ & $0.32^{0.43}_{0.20}$ & $0.22^{0.27}_{0.20}$ & $0.13^{0.14}_{0.12}$ & $0.86$ & $404^{491}_{198}$ \\
    $30$ & $0$ & $0$ & $0.0026^{0.0067}_{0}$ & $0.022^{0.043}_{0.0096}$ & $0^{0.0091}_{0}$ & $0.35^{0.44}_{0.24}$ & $0.21^{0.23}_{0.19}$ & $0.12^{0.14}_{0.11}$ & $0.83$ & $376^{532}_{218}$ \\
    $35$ & $0$ & $0$ & $0.0033^{0.0086}_{0}$ & $0.024^{0.050}_{0.011}$ & $0.0017^{0.0096}_{0}$ & $0.36^{0.45}_{0.27}$ & $0.21^{0.23}_{0.18}$ & $0.12^{0.13}_{0.11}$ & $0.83$ & $369^{566}_{199}$ \\
    $40$ & $0$ & $0$ & $0.0030^{0.0080}_{0}$ & $0.029^{0.051}_{0.0097}$ & $0^{0.0085}_{0}$ & $0.36^{0.45}_{0.29}$ & $0.20^{0.23}_{0.18}$ & $0.12^{0.13}_{0.12}$ & $0.83$ & $416^{500}_{204}$ \\
    $45$ & $0$ & $0$ & $0.0032^{0.0083}_{0}$ & $0.025^{0.049}_{0.013}$ & $0.0018^{0.0093}_{0}$ & $0.38^{0.50}_{0.27}$ & $0.21^{0.23}_{0.18}$ & $0.12^{0.14}_{0.12}$ & $0.83$ & $414^{561}_{204}$ \\
    $50$ & $0$ & $0$ & $0.0031^{0.0081}_{0}$ & $0.027^{0.044}_{0.012}$ & $0^{0.0088}_{0}$ & $0.36^{0.46}_{0.25}$ & $0.21^{0.24}_{0.18}$ & $0.12^{0.13}_{0.12}$ & $0.84$ & $428^{518}_{190}$ \\
    best & $0$ & $0$ & $0.0028^{0.0076}_{0}$ & $0.027^{0.047}_{0.011}$ & $0^{0.0085}_{0}$ & $0.33^{0.43}_{0.23}$ & $0.21^{0.23}_{0.18}$ & $0.12^{0.13}_{0.12}$ & $0.84$ & $436^{574}_{214}$ \\
    \bottomrule
  \end{tabular}%
  }
\end{table}

\begin{table}[H]
  \centering
  \footnotesize
  \caption{HierVAE local identity-support organization across training. Left: cross-neighborhood median Jaccard per equivalence convention. Right: within- and across-neighborhood ECFP Tanimoto. Each cell reports the median$^{Q_3}_{Q_1}$ over neighborhoods.}
  \label{tab:hiervae_ball_convergence}
  \resizebox{\linewidth}{!}{%
  \begin{tabular}{lcccccccccc}
    \toprule
    & \multicolumn{6}{c}{Cross-neighborhood Jaccard} & \multicolumn{4}{c}{ECFP cohesiveness} \\
    \cmidrule(lr){2-7} \cmidrule(lr){8-11}
    steps & SMILES & InChIKey-14 & murcko & murcko-gen & formula & element & med$(W)$ & med$(A)$ & AUC$_{W,A}$ & $n_{\mathrm{unique}}$ \\
    \midrule
    $5000$ & $0$ & $0$ & $0^{0.071}_{0}$ & $0.043^{0.12}_{0}$ & $0$ & $0.20^{0.29}_{0.11}$ & $0.20^{0.25}_{0.17}$ & $0.12^{0.13}_{0.10}$ & $0.78$ & $14^{20}_{8}$ \\
    $10000$ & $0$ & $0$ & $0^{0.025}_{0}$ & $0.044^{0.093}_{0}$ & $0$ & $0.22^{0.33}_{0.12}$ & $0.18^{0.22}_{0.15}$ & $0.12^{0.12}_{0.099}$ & $0.75$ & $17^{35}_{12}$ \\
    $15000$ & $0$ & $0$ & $0^{0.021}_{0}$ & $0.037^{0.086}_{0}$ & $0$ & $0.27^{0.40}_{0.17}$ & $0.21^{0.24}_{0.15}$ & $0.11^{0.12}_{0.10}$ & $0.79$ & $30^{56}_{18}$ \\
    $20000$ & $0$ & $0$ & $0^{0.029}_{0}$ & $0.056^{0.10}_{0.013}$ & $0$ & $0.30^{0.42}_{0.18}$ & $0.20^{0.22}_{0.17}$ & $0.12^{0.12}_{0.10}$ & $0.83$ & $34^{52}_{25}$ \\
    $25000$ & $0$ & $0$ & $0$ & $0.030^{0.074}_{0}$ & $0$ & $0.25^{0.40}_{0.14}$ & $0.19^{0.25}_{0.15}$ & $0.12^{0.13}_{0.11}$ & $0.81$ & $46^{71}_{32}$ \\
    $30000$ & $0$ & $0$ & $0^{0.014}_{0}$ & $0.038^{0.081}_{0.010}$ & $0$ & $0.29^{0.40}_{0.18}$ & $0.20^{0.25}_{0.18}$ & $0.12^{0.13}_{0.11}$ & $0.83$ & $50^{72}_{35}$ \\
    $50000$ & $0$ & $0$ & $0$ & $0.014^{0.042}_{0}$ & $0$ & $0.28^{0.40}_{0.18}$ & $0.21^{0.26}_{0.18}$ & $0.12^{0.12}_{0.11}$ & $0.87$ & $82^{117}_{50}$ \\
    $100000$ & $0$ & $0$ & $0^{0.0047}_{0}$ & $0.015^{0.044}_{0}$ & $0$ & $0.24^{0.36}_{0.14}$ & $0.22^{0.26}_{0.17}$ & $0.11^{0.12}_{0.10}$ & $0.87$ & $86^{132}_{61}$ \\
    $150000$ & $0$ & $0$ & $0$ & $0.013^{0.041}_{0}$ & $0$ & $0.24^{0.38}_{0.14}$ & $0.21^{0.25}_{0.17}$ & $0.11^{0.12}_{0.10}$ & $0.88$ & $80^{119}_{49}$ \\
    $200000$ & $0$ & $0$ & $0$ & $0.016^{0.044}_{0}$ & $0$ & $0.20^{0.33}_{0.12}$ & $0.22^{0.25}_{0.17}$ & $0.11^{0.12}_{0.098}$ & $0.88$ & $63^{91}_{40}$ \\
    $240000$ & $0$ & $0$ & $0^{0.013}_{0}$ & $0.019^{0.048}_{0}$ & $0$ & $0.25^{0.33}_{0.13}$ & $0.23^{0.29}_{0.17}$ & $0.12^{0.13}_{0.10}$ & $0.87$ & $58^{104}_{35}$ \\
    \bottomrule
  \end{tabular}%
  }
\end{table}
\begin{figure}[H]
    \centering
    \includegraphics[width=\linewidth]{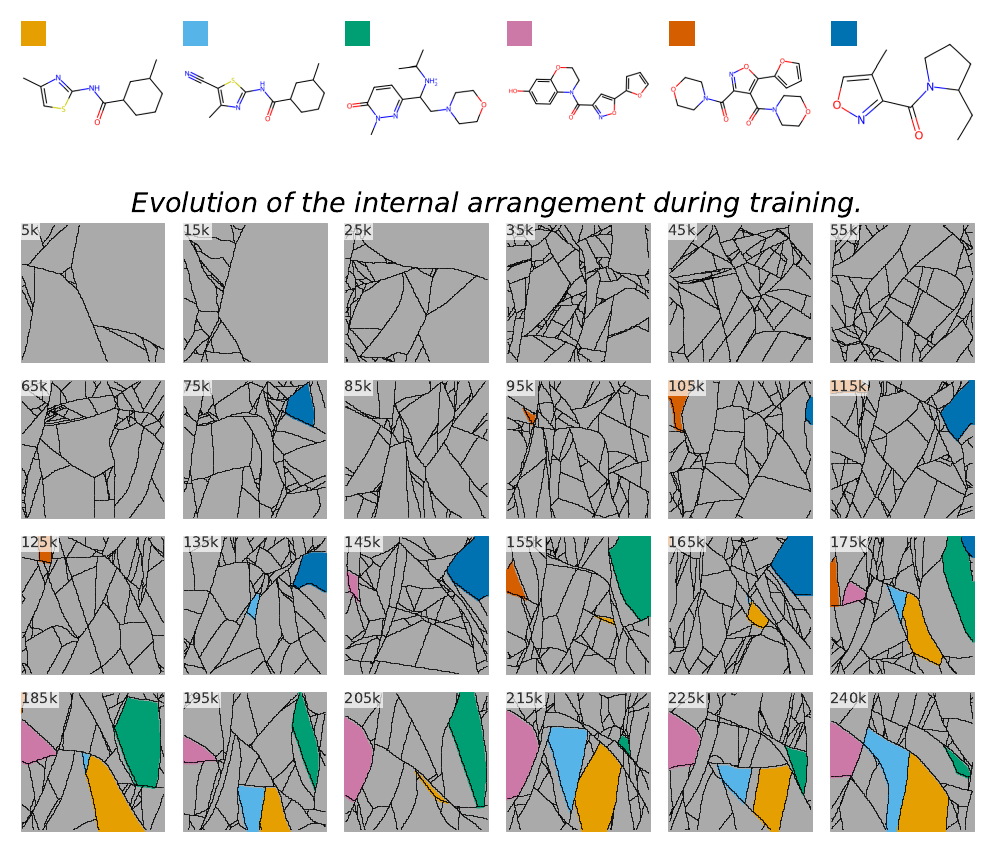}
    \caption{
    \textbf{Evolution of a HierVAE cross-section during training.}
    The same fixed two-dimensional section is decoded across different checkpoints. Regions are colored by molecular identity. The six most persistent identities across all snapshots are highlighted, while all other identities are shown in gray. Panel labels give the number of optimization steps.
    }
    \label{fig:evolution-hiervae}
\end{figure}

\section{Additional example: cross-section as a function of decoding step}
\label{sec:si-extra-lineages}

\begin{figure}[H]
    \centering
    \begin{subfigure}{\linewidth}
        \includegraphics[width=0.9\linewidth]{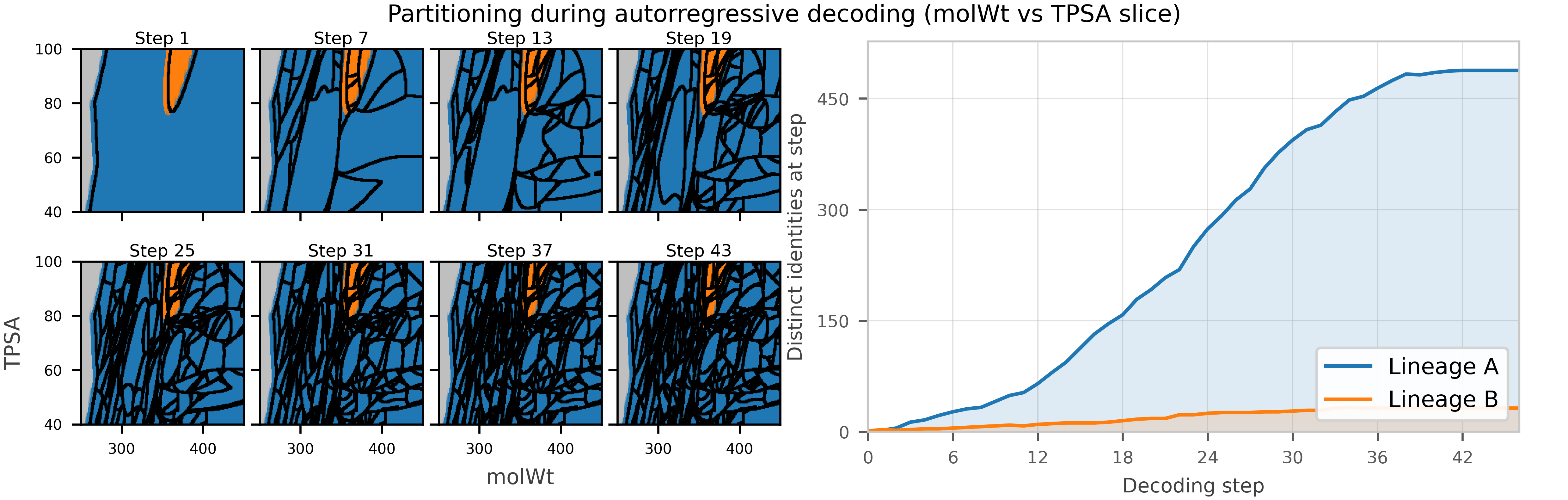}
    \end{subfigure}
    \caption{
    \textbf{Partition subdivision as a function of decoding depth (additional example).}
    Partition of a fixed MolMiner two-dimensional section as a function of decoding step. Blue and orange label cells originating from two distinct starting fragments (lineages). Right: number of distinct molecules as a function of decoding step for each lineage.
    }
    \label{fig:si-lineages}
\end{figure}

\section{One-dimensional fixed-randomness paths}\label{sec:si-1d-paths}
\begin{figure}[H]
    \centering
    \begin{subfigure}[b]{0.42\linewidth}
        \centering
        \includegraphics[width=\linewidth]{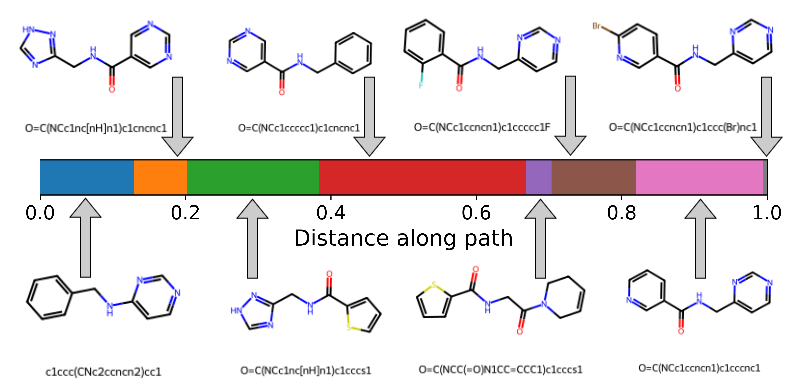}
        \caption{MolMiner}
        \label{fig:walk-molminer}
    \end{subfigure}
    \hfill
    \begin{subfigure}[b]{0.5\linewidth}
        \centering
        \includegraphics[width=\linewidth]{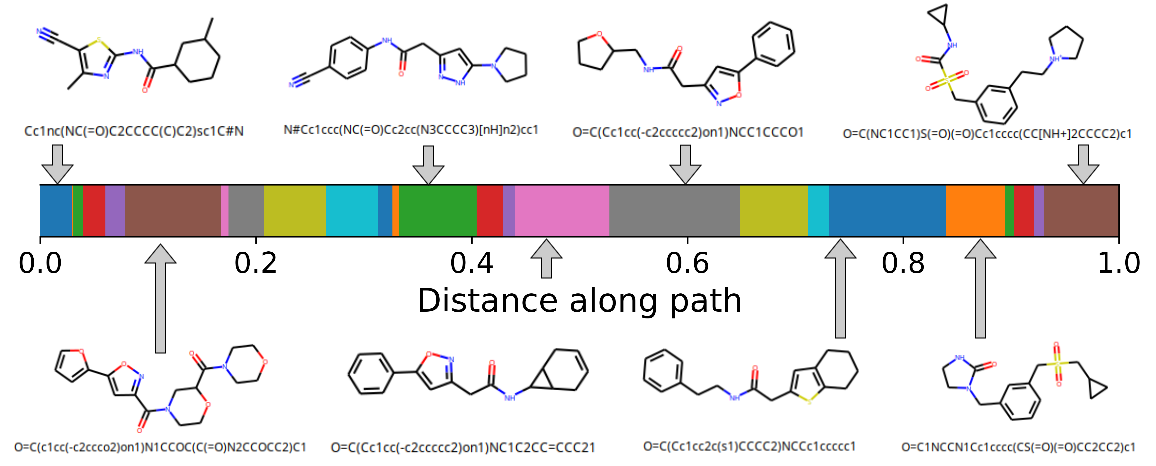}
        \caption{HierVAE}
        \label{fig:walk-hiervae}
    \end{subfigure}
    \begin{subfigure}[b]{0.45\linewidth}
        \centering
        \includegraphics[width=\linewidth]{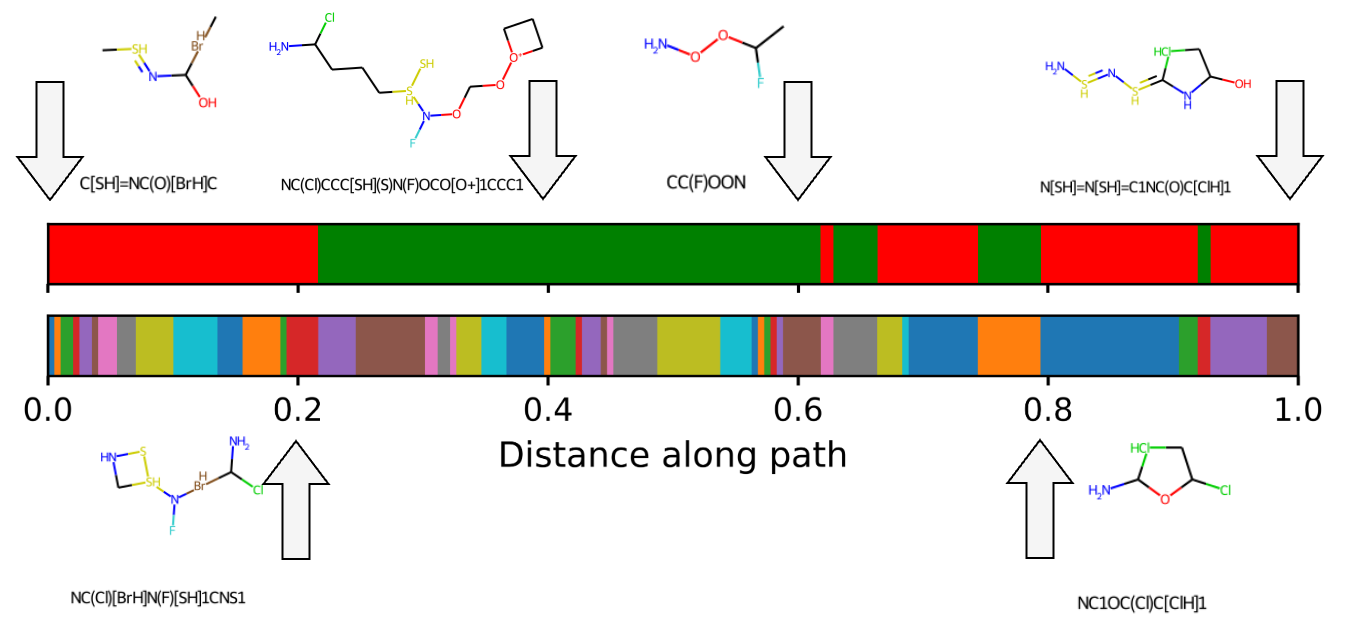}
        \caption{GDSS}
        \label{fig:walk-gdss}
    \end{subfigure}

    \caption{%
        \textbf{Fixed-randomness sections of models' partitions.}
        Decoded molecules along straight-line paths between generative coordinates. Molecular identities remain constant over intervals and switch abruptly, revealing one-dimensional intersections of identity cells. For GDSS, red segments indicate invalid outputs, corresponding to the null identity \(\varnothing\).
    }
    \label{fig:si-deterministic-tessellation}
\end{figure}

\end{document}